\documentclass[runningheads]{llncs}

\usepackage{cite}

\usepackage{float}

\newcommand{\etal}{\textit{et al}.}
\newcommand{\eg}{\textit{e.g}.,}

\usepackage{subfigure}

\usepackage{boldline,multirow}
\usepackage{tabu}
\usepackage[table]{xcolor}
\usepackage{enumitem}

\usepackage{graphicx}

\usepackage{tikz}
\usepackage{comment}
\usepackage{amsmath,amssymb} 
\usepackage{color}

\usepackage{url}

\begin{document}
\pagestyle{headings}
\mainmatter
\def\ECCVSubNumber{2293}  

\title{Self-Supervised Monocular 3D Face Reconstruction by Occlusion-Aware \\ Multi-view Geometry Consistency} 

\titlerunning{MGCNet}
%
\author{Jiaxiang Shang\inst{1}\orcidID{0000-0001-7161-9765} \and Tianwei Shen\inst{1}\orcidID{0000-0002-3290-2258} \and
		Shiwei li\inst{1}\orcidID{0000-0003-0712-0059} \and Lei Zhou\inst{1}\orcidID{0000-0003-4988-5084} \\ \and 
		Mingmin Zhen\inst{1}\orcidID{0000-0002-8180-1023} \and Tian Fang\inst{2}\orcidID{0000-0002-5871-3455} \and \\ 
		Long Quan\inst{1}\orcidID{00000001-8148-1771}}
\authorrunning{J. Shang et al.}
%
\institute{Hong Kong University of Science and Technology \email{\{jshang,tshenaa,lzhouai,mzhen,quan\}@cse.ust.hk} \\
\and Everest Innovation Technology \\ 
\email{\{sli,fangtian\}@altizure.com}}

\maketitle

\begin{abstract}
Recent learning-based approaches, in which models are trained by single-view images have shown promising results for monocular 3D face reconstruction, but they suffer from the ill-posed face pose and depth ambiguity issue. 
In contrast to previous works that only enforce 2D feature constraints, we propose a self-supervised training architecture by leveraging the multi-view geometry consistency, which provides reliable constraints on face pose and depth estimation. 
We first propose an occlusion-aware view synthesis method to apply multi-view geometry consistency to self-supervised learning. Then we design three novel loss functions for multi-view consistency, including the pixel consistency loss, the depth consistency loss, and the facial landmark-based epipolar loss.
Our method is accurate and robust, especially under large variations of expressions, poses, and illumination conditions. Comprehensive experiments on the face alignment and 3D face reconstruction benchmarks have demonstrated superiority over state-of-the-art methods. Our code and model are released in \url{https://github.com/jiaxiangshang/MGCNet}.
\end{abstract}

\section{Introduction}


3D face reconstruction is extensively studied in the computer vision community. Traditional optimization-based methods \cite{intro_relate_tra_romdhani2005estimating,intro_tra_aldrian2012inverse,intro_relate_tra_roth2015unconstrained,intro_tra_bas2016fitting, intro_tra_roth2016adaptive,intro_tra_cao2016real,intro_ava_hu2017avatar} formulate the 3D Morphable Model (3DMM) \cite{intro_3dmm_blanz1999morphable} parameters into a cost minimization problem, which is usually solved by expensive iterative nonlinear optimization.
The supervised CNN-based methods~\cite{super_syn_mul_dou2018multi,super_fit_yi2019mmface,super_fit_tran2018extreme,super_fit_volu_exp_feng2018joint,super_syn_iter_richardson20163d,super_syn_guo2018cnn,super_fiting_liu2018disentangling,super_syn_endtoend_dou2017end,super_iter_liu2016joint,super_iter_sela2017unrestricted} require abundant 3D face scans and corresponding RGB images, which are limited in amount and expensive to acquire. Methods that focus on face detail reconstruction \cite{unsuper_tran2019towards,super_chen2019photo,super_zeng2019df2net,super_review3_galteri2019deep,super_fit_tran2018extreme,super_iter_sela2017unrestricted} need even high-quality 3D faces scans.
To address the insufficiency of scanned 3D face datasets, some unsupervised or self-supervised methods are proposed \cite{unsuper_mul_ng2019accurate,unsuper_zhou2019dense,unsuper_yoon2019self,unsuper_mul_tewari2019fml,unsuper_tran2019towards,unsuper_genova2018unsupervised,unsuper_tewari2018self,unsuper_tran2018nonlinear,unsuper_tewari2017mofa,unsuper_richardson2017learning}, which employ the 2D facial \emph{landmark loss} between inferred 2D landmarks projected from 3DMM and the ground truth 2D landmarks from images, as well as the \emph{render loss} between the rendered images from 3DMM and original images. 
One critical drawback of existing unsupervised methods is that both \emph{landmark loss} and \emph{render loss} are measured in projected 2D image space and do not penalize incorrect face pose and depth value of 3DMM, resulting in the ambiguity issue of the 3DMM in the face pose and depth estimation. \\
\indent To address this issue, we resort to the multi-view geometry consistency. 
Multi-view images not only contain 2D landmarks and pixel features but also they form the multi-view geometry constraints. Such training data is publicly available and efficient to acquire (\eg videos).
Fortunately, a series of multi-view 3D reconstruction techniques named view synthesis \cite{warp_chen1993view,debevec1996modeling,warp_fitzgibbon2005image} help to formulate self-supervised learning architecture based on multi-view geometry. View synthesis is a classic task that estimates proxy 3D geometry and establishes pixel correspondences among multi-view input images. Then they generate $ N - 1 $ synthetic target view images by compositing image patches from the other $ N - 1 $ input view images. View synthesis is commonly used in Monocular Depth Estimation (MDE) task \cite{monode_zhou2017unsupervised,monode_zhan2018unsupervised,monode_mahjourian2018unsupervised,monode_casser2019depth}. However, MDE only predicts depth map and relative poses between views without inferring camera intrinsics. The geometry of MDE is incomplete as MDE loses the relationship from 3D to 2D, and they can not reconstruct a full model in scene.
MDE also suffers from erroneous penalization due to self-occlusion. \\
\indent Inspired by multi-view geometry consistency, we propose a self-supervised Multi-view Geometry Consistency based 3D Face Reconstruction framework (MGCNet). 
The workflow of MGCNet is shown in Figure~\ref{fig:pipeline}.
$ I_{t} $ is always considered to be the \textbf{target} of multi-view data.
To simplify the following formulation, we denote all $N-1$ views adjacent to the \textbf{target view} as the \textbf{source views}.
To build up the multi-view consistency in the training process via view synthesis, we first design a covisible map that stores the mask of covisible pixels for each target-source view pair to solve self-occlusion, as the large and extreme face pose cases is common in the real world, and the self-occlusion always happens in such profile face pose cases. 
Secondly, we feed the 3DMM coefficients and face poses to the differentiable rendering module \cite{unsuper_genova2018unsupervised}, producing the rendered image, depth map, and covisible map for each view. 
Thirdly, pixel consistency loss and depth consistency loss are formulated by input images and rendered depth maps in covisible regions, which ensures the consistency of 3DMM parameters in the multi-view training process. 
Finally, we introduce the facial epipolar loss, which formulates the epipolar error of 2D facial landmarks via the relative pose of two views,
as facial landmarks is robust to illumination changes, scale ambiguity, and calibration errors.
With these multi-view supervised losses, we are able to achieve accurate 3D face reconstruction and face alignment result on multiple datasets \cite{dataset_aflw20003D_300WLP_zhu2016face,dataset_florence,dataset_bu3dfe_yin20063d,dataset_bu4dfe_yin20063d,dataset_frgc}. 
We conduct ablation experiments to validate the effectiveness of covisible map and multi-view supervised losses.
To summarize, this paper makes the following main contributions:
\begin{itemize}
\setlength{\itemsep}{0pt}
\setlength{\parsep}{0pt}
\setlength{\parskip}{0pt}
\item[-] We propose an end-to-end self-supervised architecture MGCNet for face alignment and monocular 3D face reconstruction tasks. To our best knowledge, we are the first to leverage multi-view geometry consistency to mitigate the ambiguity from monocular face pose estimation and depth reconstruction in the training process.
\item[-] We build a differentiable covisible map for general view synthesis, which can mitigate the self-occlusion crux of view synthesis. Based on view synthesis, three differentiable multi-view geometry consistency loss functions are proposed as pixel consistency loss, depth consistency loss, and facial epipolar loss.
\item[-] Our MGCNet result on the face alignment benchmark \cite{dataset_aflw20003D_300WLP_zhu2016face} shows that we achieve more than a $ \mathbf{12\%} $ improvement over other state-of-the-art methods, especially in large and extreme face pose cases. Comparison on the challenging 3D Face Reconstruction datasets \cite{dataset_florence,dataset_bu3dfe_yin20063d,dataset_bu4dfe_yin20063d,dataset_frgc} shows that MGCNet outperforms the other methods with the largest margin of $ \mathbf{17\%} $.
\end{itemize}

\section{Related work}
\subsection{Single-view Method} 

Recent CNN methods
\cite{dataset_aflw20003D_300WLP_zhu2016face,super_fit_volu_jackson2017large,super_fit_endtoend_iter_tuan2017regressing,super_fiting_liu2018disentangling,super_fit_tran2018extreme,super_fit_volu_exp_feng2018joint,super_fit_yi2019mmface,unsuper_tran2019towards,super_chen2019photo,super_zeng2019df2net,super_review3_galteri2019deep,super_syn_endtoend_dou2017end,super_syn_guo2018cnn} train the CNN network supervised by 3D face scan ground truth and achieve impressive results. 
\cite{super_syn_iter_richardson20163d,super_syn_endtoend_dou2017end,super_syn_guo2018cnn} generate synthetic rendered face images with real 3D scans.
\cite{super_fit_volu_jackson2017large,super_fit_endtoend_iter_tuan2017regressing,super_fiting_liu2018disentangling,super_fit_tran2018extreme,super_fit_volu_exp_feng2018joint,super_fit_yi2019mmface} propose their deep neural networks trained using fitted 3D shapes by traditional methods as substitute labels.
%
%
%
Lack of realistic training data is still a great hindrance.

Recently, some self-supervised or weak-supervised methods are proposed \cite{unsuper_richardson2017learning,unsuper_tewari2017mofa,unsuper_tran2018nonlinear,unsuper_tewari2018self,unsuper_genova2018unsupervised,unsuper_mul_sanyal2019learning_ring,unsuper_tran2019towards,unsuper_yoon2019self,unsuper_zhou2019dense} to solve the lack of high-quality 3D face scans with robust testing result.
Tewari \etal \cite{unsuper_tewari2017mofa} propose an differentiable rendering process to build unsupervised face autoencoder based on pixel loss.
Genova \etal \cite{unsuper_genova2018unsupervised} train a regression network mainly focus on identity loss that compares the features of the predicted face and the input photograph. 
Nevertheless, face pose and depth ambiguity originated from only monocular images still a limitation.

\subsection{Multi-view or Video Based Method} 

There are established toolchains of 3D reconstruction \cite{sfm_furukawa2010towards,sfm_wu2011visualsfm,sfm_schoenberger2016sfm,mvs_schoenberger2016mvs}, aimming at recovering 3D geometry from multi-view images.
One related operation is view synthesis \cite{warp_chen1993view,warp_seitz1996view,warp_fitzgibbon2005image}, and the goal is to synthesize the appearance of the scene from novel camera viewpoints.

Several unsupervised approaches \cite{unsuper_mul_ng2019accurate,unsuper_mul_wu2019mvf,unsuper_mul_tewari2019fml,unsuper_mul_sanyal2019learning_ring} are proposed recently to address the 3D face reconstruction from multiple images or videos.
Deng \etal \cite{unsuper_mul_ng2019accurate} perform multi-image face reconstruction from different images by shape aggregation.
Sanyal \etal \cite{unsuper_mul_sanyal2019learning_ring} take multiple images of the same and different person, then enforce shape consistency between the same subjects and shape inconsistency between the different subjects.
Wu \etal \cite{unsuper_mul_wu2019mvf} design an impressive multi-view framework (MVFNet), which is view-consistent by design, and photometric consistency is used to generate consistent texture across views. However, MVFNet is not able to generate results via a single input since it relies on multi-view aggregation during inference.
Our MGCNet explicitly exploit multi-view consistency (both geometric and photometric) to constrain the network to produce view-consistent face geometry from \textbf{a single input}, which provides better supervision than 2D information only. Therefore, MGCNet improves the performance of face alignment and 3D face reconstruction as Section \ref{sec:exp}.

\begin{figure*}[t]
	\centering
	\includegraphics[width=0.95\linewidth]{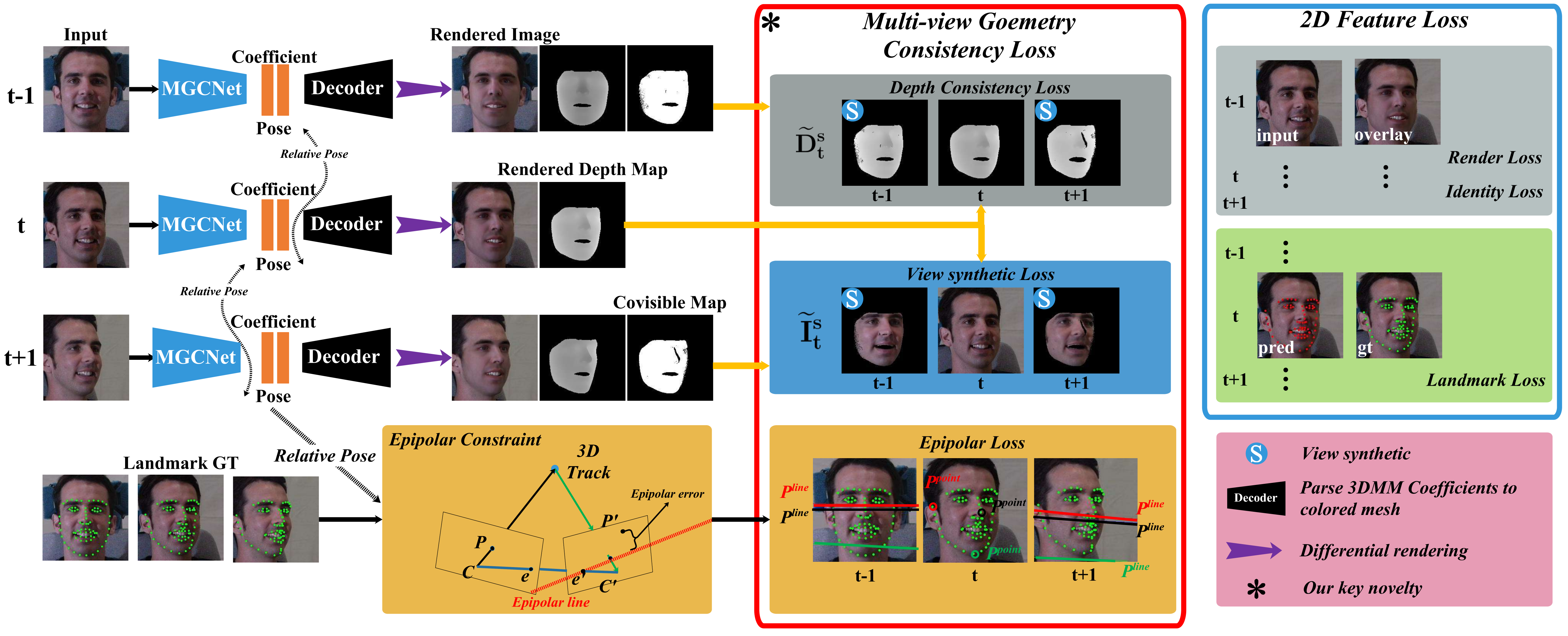}
	\caption{The training flow of our MGCNet architecture, which is annotated in Section \ref{sec:method_overview}. The 2D feature loss part is our \textbf{baseline} in Section \ref{sec:method_common_loss}.
		Our novel multi-view geometry consistency loss functions are highlighted as $*$ in Section \ref{sec:method}. 
	}
	\label{fig:pipeline}
\end{figure*}

\section{Method} \label{sec:method}

\subsection{Overview} \label{sec:method_overview}
The inference process of MGCNet takes a single image as input, while in the training process the input is $N$-view images (e.g. $ {I_{t-1}, I_{t}, I_{t+1}} $ for $N = 3$ ) of the same face and the corresponding ground-truth 2D landmarks $ q_{t-1}^{gt}, q_{t}^{gt}, q_{t+1}^{gt} $.
Then, the MGCNet estimates the 3DMM coefficients and face poses, whose notations are introduced in Section \ref{method_model} in detail. $\widetilde{\mathcal{I}}_t^{(s)}, \widetilde{\mathcal{D}}_t^{(s)}$ represent the synthesized target images and the depth maps from the source views.

%

\subsection{Model} \label{method_model}
\textbf{Face model} 3D Morphable Model (3DMM) proposed by \cite{intro_3dmm_blanz1999morphable} is the face prior model of the MGCNet.
Specifically, the 3DMM encodes both face shape and texture as 
\begin{equation}
\label{eq:3dmm_model}
\begin{split}
\small
& S = S(\alpha,\beta) = S_{mean} + A_{id}\alpha + B_{exp}\beta \\
& T = T(\gamma) = T_{mean} + T_{id}\gamma, \\
\end{split}
\end{equation}
where $ S_{mean} $ and $ T_{mean} $ denote the mean shape and the mean albedo respectively.
$ A_{id} $, $ B_{exp} $ and $ T_{id} $ are the PCA bases of identity, expression and texture.
$ \alpha, \beta \in R^{80} $ and $ \gamma \in R^{64} $ are corresponding coefficient vectors to be estimated follow \cite{unsuper_mul_ng2019accurate,unsuper_tewari2017mofa,unsuper_tewari2018self}.
We use $ S_{mean} $, $ T_{mean} $, $ A_{id} $ and $ T_{id} $ provided by the Basel Face Model (BFM) \cite{intro_3dmm_blanz1999morphable,3dmm_paysan20093d}, and $ B_{exp} $ from FaceWarehouse \cite{dataset_cao2013facewarehouse}. 
We exclude the ear and neck region as \cite{unsuper_mul_ng2019accurate}, and our final face model contains $ \sim36K $ vertices. 

\noindent
\textbf{Camera model} 
The pinhole camera model is employed to define the 3D-2D projection.
We assume the camera is calibrated.
The face pose $ P $ is represented by an euler angle rotation $ R \in SO(3)$ and translation $ t \in R^{3} $.
The relative poses $ P^{rel}_{t \rightarrow s} \in SE(3) $ from the \textbf{target view} to $ N - 1 $ \textbf{source views} are defined as $ P^{rel}_{t \rightarrow s} = 
\left[ \begin{array}{cc}
R_{t}^{-1} R_{s} & R_{t}^{-1}(t_{s}-t_{t}) \\
0 & 1 
\end{array} \right]$.


\noindent
\textbf{Illumination model}
To acquire realistic rendered face images, we model the scene illumination by Spherical Harmonics (SH)  \cite{sh_ramamoorthi2001efficient,sh_ramamoorthi2001signal} as $  SH(N_{re},T|\theta) = T * \sum_{b=1}^{B^{2}}\theta_{b} H_{b} $,
where $N_{re}$ is the normal of the face mesh, $ \theta \in R^{27} $ is the coefficient. 
The $ H_{b} : R^{3} \rightarrow R $ are SH basis functions and the $ B^{2} = 9 $ (B = 3 bands) parameterizes the colored illumination in red, green and blue channels.

Finally, we concatenate all the parameters together into a $ (\alpha, \beta, \gamma, R, t, \theta) $ 257-dimensional vector. 
All 257 parameters encode the 3DMM coefficients and the face pose, which are abbreviated as \textit{coefficient} and \textit{pose} in Figure \ref{fig:pipeline}.

\subsection{2D Feature Loss} \label{sec:method_common_loss}
Inspired by recent related works, we leverage preliminary 2D feature loss functions in our framework.

\noindent
\textbf{Render loss} 
The render loss aims to minimize the difference between the input face image and the rendered image as $ \mathcal L_{render} = \frac{1}{M} \sum_{i=1}^{M} w_{skin}^{i} || I^{i} - I_{re}^{i} || $,
where $ I_{re} $ is the rendered image, $ I $ is the input image, and M is the number of all 2D pixels in the projected 3D face region. 
$ w_{skin}^{i} $ is the skin confidence of $i^{th}$ pixel as in \cite{unsuper_mul_ng2019accurate}. Render loss mainly contributes to the albedo of 3DMM.

\noindent
\textbf{Landmark loss} 
To improve the accuracy of face alignment, we employ the 2D landmark loss which defines the distance between predicted landmarks and ground truth landmarks as $  \mathcal L_{lm} = \sum_{i=1}^{N} c_{lm}^{i} (q_{gt}^{i} - q^{i})^{2}, $
where $ N $ is the number of landmarks, and $q$ the projection of the 3D landmarks picked from our face model.
It is noted that the landmarks have different levels of importance, denoted as the confidence $ c_{lm} $ for each landmark. We set the confidence to 10 only for the nose and inner mouth landmarks, and to 1 else wise.

\noindent
\textbf{Identity loss} 
The fidelity of the reconstructed face is an important criterion.
We use the identity loss as in \cite{unsuper_genova2018unsupervised}, which is the cosine distance between deep features of the input images and rendered images as $ \mathcal L_{id} = \frac{\eta_{1} \circ \eta_{2}}{|\eta_{1}| |\eta_{2}|} $, where $ \circ $ means the element-wise multiplication. $ \eta_{1} $ and $ \eta_{2} $ are deep features of input images and rendered images. 

\noindent
\textbf{Regularization loss} 
To prevent the face shape and texture parameters from diverging, regularization loss of 3DMM is used as $ \mathcal L_{reg} = w_{id} \sum_{i=1}^{N_{\alpha}}\alpha^{2} + w_{exp} \sum_{i=1}^{N_{\beta}}\beta^{2} + w_{tex} \sum_{i=1}^{N_{\gamma}}\gamma^{2} $
where $ w_{id}, w_{exp}, w_{shape} $ are trade-off parameters for 3DMM coefficients regularization ($ 1.0, 0.8, 3\mathrm{e}{-3} $ by default).
$ N_{\alpha}, N_{\beta}, N_{\gamma} $ are the length of 3DMM parameters $ \alpha $, $ \beta $, $ \gamma $.

\noindent
\textbf{Final 2D feature loss} 
The combined 2D feature loss function $ \mathcal L_{2D} $ is defined as $ \mathcal L_{2D} = w_{render} \mathcal L_{render} + w_{lm} \mathcal L_{lm} + w_{id} \mathcal L_{id} + w_{reg} \mathcal L_{reg}, $
where the trade-off parameters for 2D feature losses are set empirically $w_{render} = 1.9, w_{lm} = 1\mathrm{e}{-3}, w_{id} = 0.2, w_{reg} = 1\mathrm{e}{-4}$.
We regard the \textbf{baseline} approach in the later experiement as the model trained by only 2D feature losses. In the followings, we present key ingredients of our contributions.

\subsection{Occlusion-Aware View Synthesis} \label{sec:oavs}
\begin{figure}[htbp]
	\centering  
	\subfigure[]{
		\label{fig:eg_ob_occlu.sub.1}
		\includegraphics[width=0.15\textwidth]{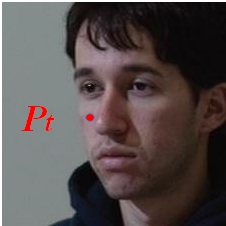}}
	\subfigure[]{
		\label{fig:eg_ob_occlu.sub.2}
		\includegraphics[width=0.15\textwidth]{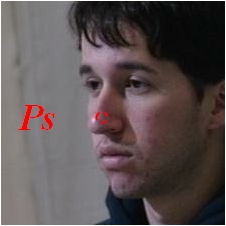}}
	\subfigure[]{
		\label{fig:eg_ob_occlu.sub.bcm}
		\includegraphics[width=0.15\textwidth]{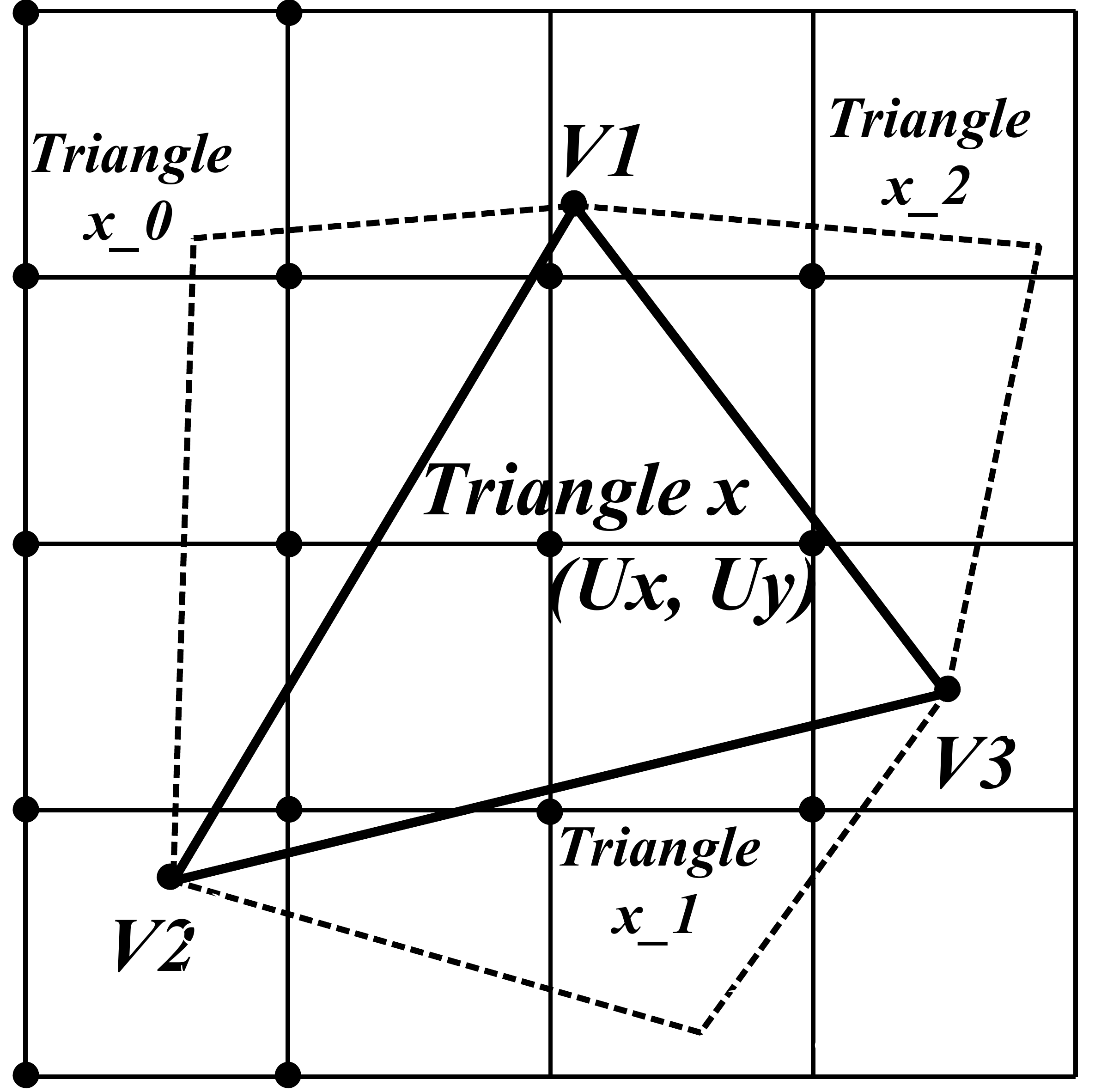}}
	\subfigure[]{
		\label{fig:eg_ob_occlu_covis}
		\includegraphics[width=0.15\textwidth]{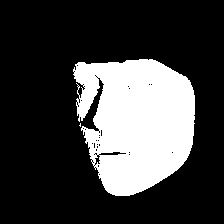}}
	\subfigure[]{
		\label{fig:eg_ob_occlu.sub.4}
		\includegraphics[width=0.15\textwidth]{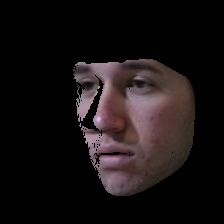}}
	
	\caption{(a) and (b) are the target view and source view pair; (c) is the covisible points and triangles; (d) is the covisible map; and (e) is the synthetic target view}
	\label{fig:eg_ob_occlu_bcm}
\end{figure}

\begin{figure}[htbp]
	\centering
	\includegraphics[width=0.97\linewidth,scale=1.00]{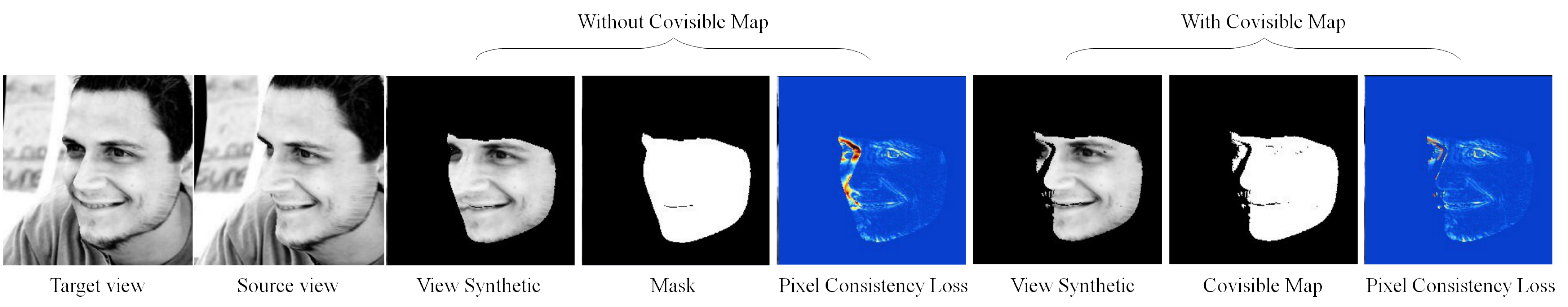}
	\caption{The view synthesis results with and without covisible map. Without covisiable map, the pixel consistency loss is highly affected by self-occlusion.
	}
	\label{fig:train_cov_right}
\end{figure}

The key of our idea is to enforce multi-view geometry consistency, so as to achieve the self-supervised training.
This could be done via view synthesis, which establishes dense pixel correspondences across multi-view input images. However, the view synthesis for face reconstruction is very easily affected by self-occlusion, as large and extreme face pose cases are common in read world applications.
As shown in Figure \ref{fig:eg_ob_occlu_bcm}, assuming a pixel $ p_{t} $ is visible in the left cheek as shown in Figure \ref{fig:eg_ob_occlu.sub.1}, the correspondence pixel  $ p_{s} $ could not be found in Figure \ref{fig:eg_ob_occlu.sub.2} due to nose occlusion.
Self-occlusion leads to redundant pixel consistency loss and depth consistency loss. Furthermore the related gradient of self-occlusion pixels will be highly affected by the salient redundant error as red part in Figure \ref{fig:train_cov_right} (Pixel Consistency Loss subfigure), which makes the training more difficult.
For more practical and useful navigation in real scenarios, self-occlusion is worth to solve.

We introduce the covisible maps $ C_{s} $ to account for the self-occlusion. 
Covisible map is a binary mask indicating the pixels which are visible in both source and target views. 
During the rendering process of the MGCNet, rasterization builds the correspondence between vertices of a triangle and image pixels $ (V_{1,2,3} \sim U_{x}, U_{y}) $, as shown in Figure \ref{fig:eg_ob_occlu.sub.bcm}. 
The common vertices visible in two views (i.e., vertices that contribute to pixel rendering) are called covisible points. Then we define all triangles adjacent to covisible points as covisible triangles.
Finally, we project covisible triangles of the 3D face from the target view to image space, as shown in Figure \ref{fig:eg_ob_occlu_covis}, where the white region is covisible region. The improvement brings from covisible maps is elaborated in Figure \ref{fig:train_cov_right}, pixels are not covisible in the left of the nose in target view (red in Figure \ref{fig:train_cov_right}), which result in redundant error. The quantitative improvements are discussed in Section \ref{sec_abla}.

To generate the synthetic target RGB images from source RGB images, we first formulate the pixel correspondences between view pairs $ (I_{s}, I_{t}) $.
Given a pair correspondence pixel coordinate $ p_{t}, p_{s}$ in $ I_{t}, I_{s} $, the pixel value $ p_{s} $ is computed by bilinear-sampling \cite{jaderberg2015spatial}, and the pixel coordinate $ p_{s} $ is defined as
\begin{equation}\label{eq:warp_pixel}
\small
p_{s} \sim \mathbf{K}_s [\mathbf{P}_{t \rightarrow s}^{rel}] \mathbf{D}_t(p_t)\mathbf{K}_t^{-1} p_t,
\end{equation}
where $\sim$ represents the equality in the homogeneous coordinates, $\mathbf{K}_s$ and $\mathbf{K}_t$ are the intrinsics for the input image pairs, $\mathcal{D}_t$ is the rendered depth map of the target view, and $ \mathbf{D}_t(p_t)$ is the depth for this particular pixel $p_t$ in $\mathcal{D}_t$.

\subsection{Pixel Consistency Loss} \label{sec:pixel}
We generate the synthesized target images by view synthesis, then we minimize the pixel error between the target view and the synthesized target views from the source views as
\begin{equation}\label{eq:loss_pixel}
\small
\mathcal L_{pixel} = \frac{1}{|C_{s}|}\sum_{i=1}^{|C_{s}|} C_{s}^{i} * \left| \widetilde{ \mathcal{I}}_t^{s}(i) - \mathcal{I}_t(i) \right|,
\end{equation}
where $\widetilde{\mathcal{I}}_t^{s}$ represents the synthesized target views from the source views.
$\mathcal{I}_{t}(i)$ is the $ i-th $ pixel value .
Concretely, the first term $\widetilde{\mathcal{I}}_t^{s}$ is the bilinear-sampling operation, which computes the corresponding pixel coordinates using the relative pose $ P^{rel}_{t \rightarrow s} $ and target depth map $ D_{t}^{re}$. $C_{s}$ is covisible map and $|C_{s}|$ denotes the total number of covisible pixels.
%

\subsection{Dense Depth Consistency Loss} \label{sec:depth}
Compared to RGB images, 
depth maps are less adversely affected by the gradient locality issue \cite{bergen1992hierarchical}.
Thus, we propose a dense depth consistency loss function which contributes to solving depth ambiguity more explicitly, enforcing the multi-view consistency upon depth maps.
Similarly, we synthesize the target depth maps $\widetilde{\mathbf{D}}_t^{s}$ from the source views via bilinear interpolation, and compute the consistency against the target depth map $\mathbf{D}_t$.

One critical issue is that the face region is cropped in the face detection data-preprocessing stage, making the depth value up to scale. 
To tackle this issue, we compute a ratio of two depth maps $ \mathcal{S}_{depth} $ and rectify the scale of depth maps. Therefore, we define the dense depth consistency loss as
\begin{equation}\label{eq:depth_loss}
\small
\begin{split}
\mathcal{S}_{depth} & = \frac{\sum_{i=1}^{|C_{s}|}\mathbf{D}_t(i) C_{s}(i)}{\sum_{i=1}^{|C_{s}|}\widetilde{\mathbf{D}}_t^{s}(i) C_{s}(i)} \\
\mathcal{L}_{depth} & = \frac{1}{|C_{s}|}\sum_{i=1}^{|C_{s}|} \left| \mathcal{S}_{depth} \cdot \widetilde{\mathbf{D}}_t^{s}(i) - \mathbf{D}_t(i)\right|,
\end{split}
\end{equation}
where $\mathcal{C}_{s}(i), \mathcal{D}_{t}(i)$ are the $i-th$ covisible and depth value.
$\mathcal{S}_{depth}$ is the depth scale ratio.
Our experiment shows that the multi-view geometry supervisory signals significantly improve the accuracy of the 3D face shape.

\begin{figure}[htbp]
	\centering  
	\subfigure[Baseline]{
		\label{fig:ab_epi.sub.0}
		\includegraphics[width=0.42\textwidth]{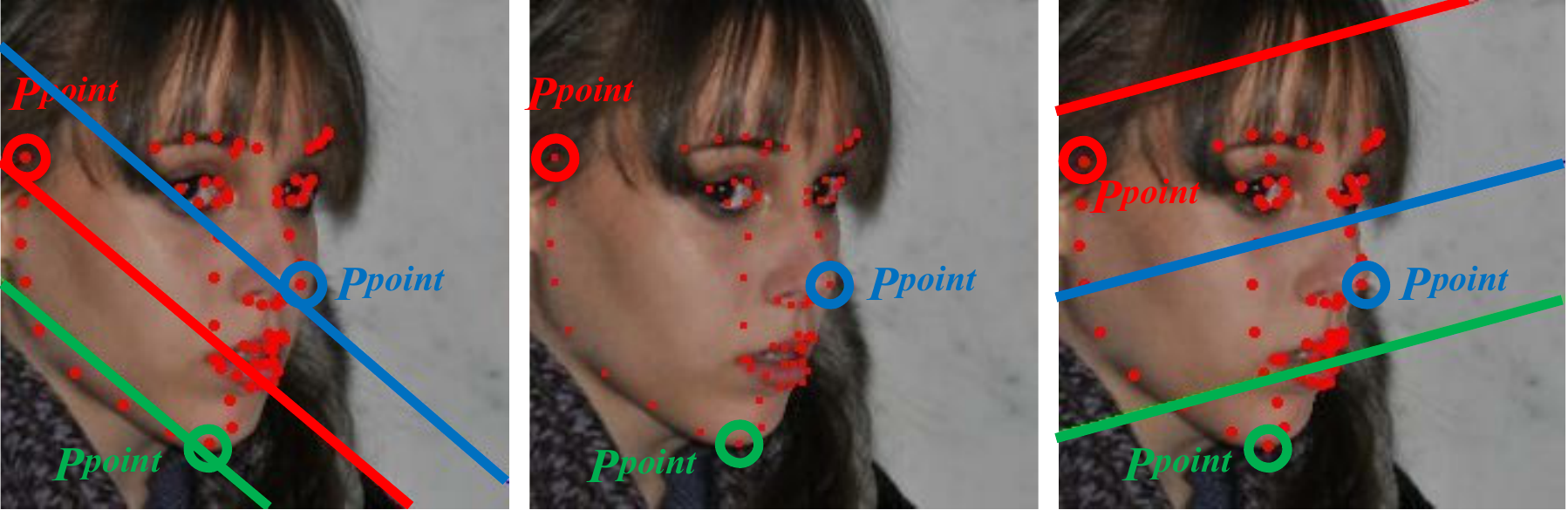}}
	\subfigure[Multi-view]{
		\label{fig:ab_epi.sub.1}
		\includegraphics[width=0.42\textwidth]{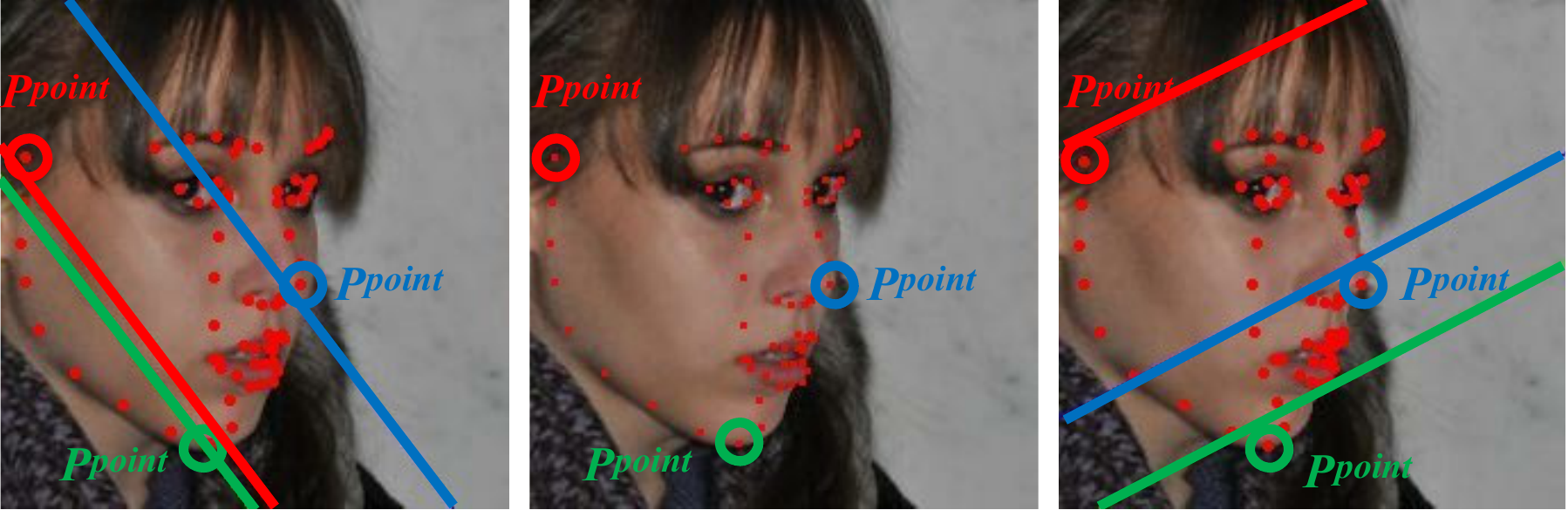}}
	\caption{Eipipolar error from target to source views, (a) is the epipolar visualization from baseline network trained by 2D feature loss; and (b) is epipolar visualization of the MGCNet. The red, green and blue point means left ear bound, lower jaw and nose tip landmarks. The epipolar error of baseline is significantly larger than MGCNet.
	}
	\label{fig:ab_epi}
\end{figure}

\subsection{Facial epipolar Loss} \label{sec:epi}
We use facial landmarks to build the epipolar consistency, as our epipolar loss is based on sparse ground-truth 2D facial landmarks, which is less likely to be affected by radiometric or illumination changes compared to pixel consistency or depth consistency losses, 
The epipolar loss in the \textit{symmetric epipolar distance}~\cite{hartley2003multiple} form between source and target 2D landmark $q_{t \leftrightarrow s} = \{\mathbf{p}\leftrightarrow \mathbf{p'}\}$ is defined as
\begin{equation} \label{eq:epi_loss}
\small
\begin{split}
\mathcal{L}_{epi}( q |\mathbf{R},\mathbf{t} )  = \sum_{\forall (\mathbf{p}, \mathbf{p'}) \in q} & \frac{\mathbf{p'^TEp}}{\sqrt{\mathbf{(E p)^2_{(1)}}+\mathbf{(E p)^2_{(2)}}}} \\ 
\end{split}
\end{equation}
where $\mathbf{E}$ being the essential matrix computed by $\mathbf{E} = [\mathbf{t}]_\times \mathbf{R}$, $[\cdot]_\times$ is the matrix representation of the cross product with $\mathbf{t}$. We simply omit the subindices for conciseness ($q$ for $q_{t \leftrightarrow s}$, $\mathbf{R}$ for $\mathbf{R}_{t \rightarrow s}$, $\mathbf{t}$ for $\mathbf{t}_{t \rightarrow s}$).


\subsection{Combined Loss}
The final loss function $ \mathcal{L} $ for our MGCNet is the combination of 2D feature loss and multi-view geometry loss.
Training the network by only 2D feature losses leads to face pose and depth ambiguity, which is reflected in geometry inconsistency as the large epipolar error shown in Figure \ref{fig:ab_epi.sub.0}.
Our MGCNet trained with pixel consistency loss, dense depth consistency, and facial epipolar loss shows remarkable improvement in Figure \ref{fig:ab_epi.sub.1}, which outstands our novel multi-view geometry consistency based self-supervised training pipeline.
Finally, the combined loss function is defined as
\begin{equation} \label{eq:loss_all}
\small
\begin{split}
\mathcal{L} = & w_{2D} * \mathcal L_{2D} + w_{mul} * [ w_{pixel} * \mathcal L_{pixel} + w_{depth} * \mathcal L_{depth} + w_{epi} * \mathcal L_{epi}],              
\end{split}
\end{equation}
where $w_{2D}=1.0$ and $w_{mul}=1.0$ balance the weights between the 2D feature loss for each view and the multi-view geometry consistency loss. 
The trade-off parameters to take into account are $ w_{pixel} = 0.15, w_{depth} = 1\mathrm{e}{-4}, w_{epi}=1\mathrm{e}{-3}$.
\section{Experiment} \label{sec:exp}
We evaluate the performance of our MGCNet on the face alignment and 3D face reconstruction tasks which compared with the most recent state-of-the-art methods \cite{unsuper_zhou2019dense,dataset_bulat2017far_lsw3d,super_fit_stn_bhagavatula2017faster,dataset_aflw20003D_300WLP_zhu2016face, unsuper_mul_sanyal2019learning_ring, super_fit_volu_exp_feng2018joint,unsuper_genova2018unsupervised,super_fit_endtoend_iter_tuan2017regressing,unsuper_mul_ng2019accurate,unsuper_tewari2017mofa,unsuper_mul_tewari2019fml,unsuper_tewari2018self} on diverse test datasets including AFLW20003D \cite{dataset_aflw20003D_300WLP_zhu2016face}, MICC Florence \cite{dataset_florence}, Binghamton University 3D Facial Expression (BU-3DFE) \cite{dataset_bu3dfe_yin20063d,dataset_bu4dfe_yin20063d}, and FRGC v2.0 \cite{dataset_frgc}.

\subsection{Implementation Details}
\textbf{Data}
300W-LP \cite{dataset_aflw20003D_300WLP_zhu2016face} has multi-view face images with fitted 3DMM model, the model is widely used as ground truth in \cite{super_fit_endtoend_iter_tuan2017regressing,super_fit_volu_exp_feng2018joint,super_fit_volu_jackson2017large,super_fit_yi2019mmface}, such multi-view images provide better supervision than only 2D features. Multi-PIE \cite{dataset_gross2010multi} are introduced to provide multi-view face images that help solve face pose and depth ambiguity.
As multi-view face datasets are always captured indoor, and thus cannot provide diversified illumination and background for training, CelebA \cite{dataset_liu2015deep_celeba} and LS3D \cite{dataset_bulat2017far_lsw3d} are used as part of training data, which only contribute to 2D feature losses. Detail data process can be found in the \textit{suppl.} material.

\noindent
\textbf{Network}
We use the ResNet50 \cite{he2016deep} network as the backbone of our MGCNet, we only convert the last fully-connected layer to 257 neurons to match the dimension of 3DMM coefficients.
The pre-trained model from ImageNet \cite{russakovsky2015imagenet} is used as an initialization. We only use $ N=3 $ views in practice, as $ N=5 $ views lead to a large pose gap between the first view and the last view. We implement our approach by Tensorflow \cite{abadi2016tensorflow}. 
The training process is based on Adam optimizer \cite{kingma2014adam} with a batch size of 5.
The learning rate is set to $ 1\mathrm{e}{-4} $, and there are 400K total iterations for the whole training process.

%
%
%
%

\begin{figure}[htbp]
	\centering
	\includegraphics[width=0.95\linewidth,scale=1.00]{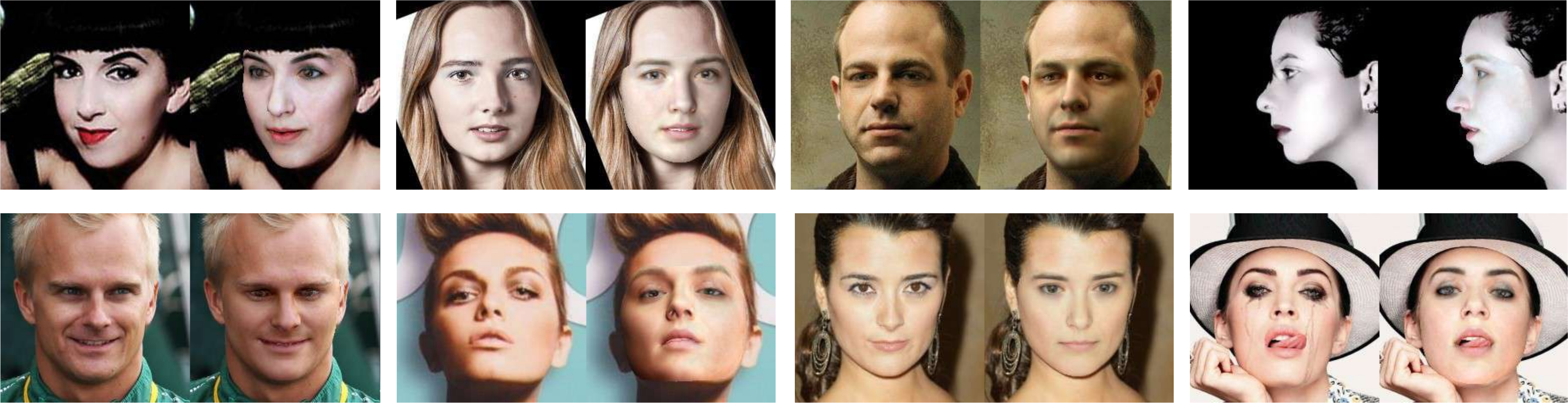}
	\caption{A few results on a full range of lighting, pose, including large expressions. Each image pair is input image (left) and reconstruction result overlay (right). Further detail result (shape, albedo, and lighting) can be found in the \textit{suppl.} material.}
	\label{fig:selfshow_flat}
\end{figure}

\subsection{Qualitative Result}

\textbf{Result in different situations}
Our MGCNet allows for high-quality reconstruction of facial geometry, reflectance and incident illumination as Figure \ref{fig:selfshow_flat}, under full range of lighting, pose, and expressions situations.

\begin{figure}[htbp]
	\centering
	\includegraphics[width=0.96\linewidth,scale=1.00]{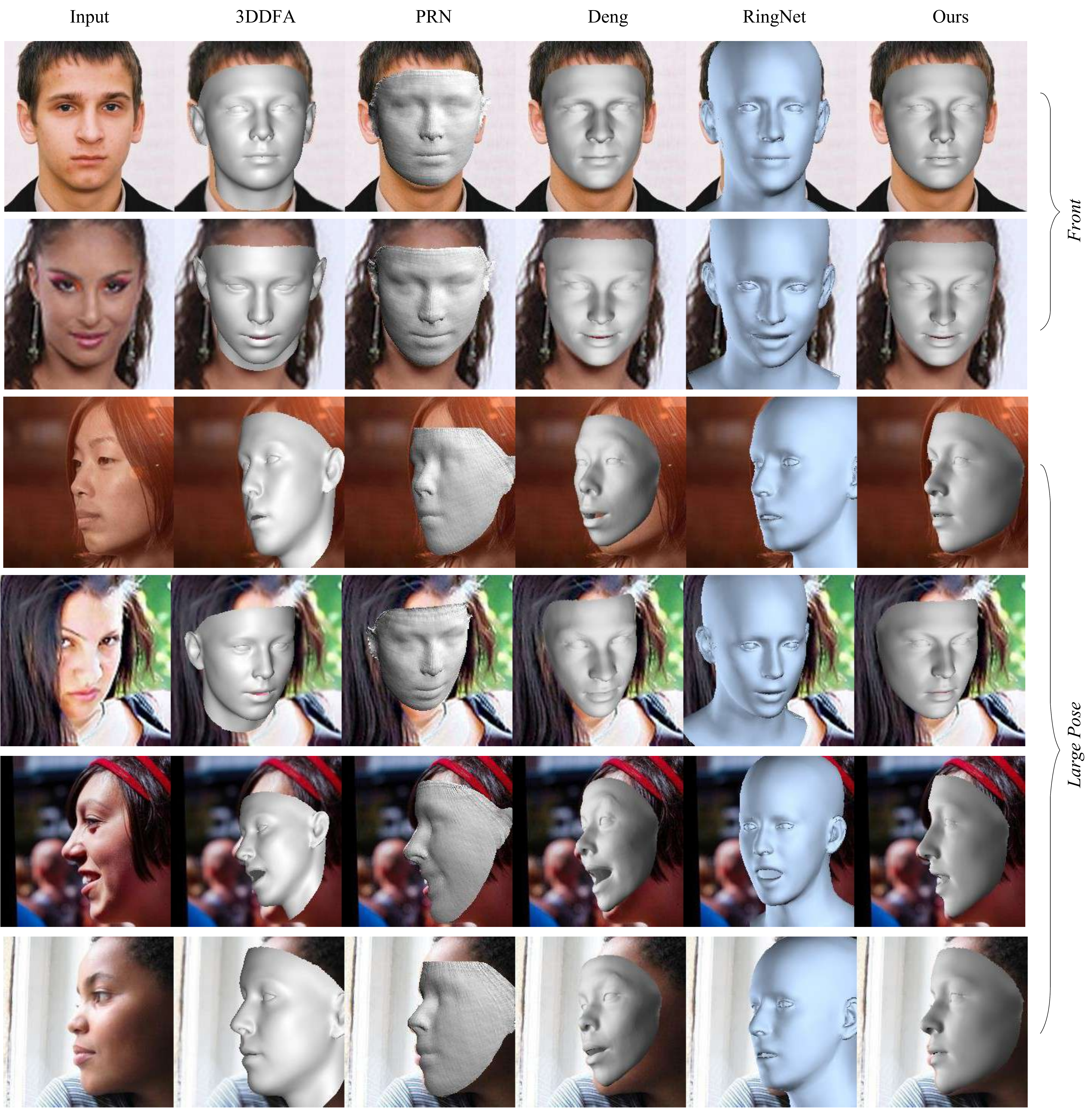}
	\caption{Comparisons with 3DDFA \cite{dataset_aflw20003D_300WLP_zhu2016face}, RingNet \cite{unsuper_mul_sanyal2019learning_ring}, PRN \cite{super_fit_volu_exp_feng2018joint}, and Deng \etal \cite{unsuper_mul_ng2019accurate} on ALFW20003D.}
	\label{fig:alfw_large}
\end{figure}

\noindent
\textbf{Geometry} We evaluate the qualitative results of our MGCNet on AFLW20003D \cite{dataset_aflw20003D_300WLP_zhu2016face}. 
First, we compare our MGCNet with 3DDFA \cite{dataset_aflw20003D_300WLP_zhu2016face}, RingNet \cite{unsuper_mul_sanyal2019learning_ring}, PRN \cite{super_fit_volu_exp_feng2018joint}, and Deng \etal \cite{unsuper_mul_ng2019accurate} on front view samples, as Row 1 and Row 2 in Figure \ref{fig:alfw_large}.
Our predicted 3DMM coefficients produce more accurate results than the most methods, and we get comparable results with Deng \etal \cite{unsuper_mul_ng2019accurate}. 

For these large and extreme pose cases as Row 3-6 in Figure \ref{fig:alfw_large}, our MGCNet has better face alignment and face geometry than other methods. 
We have more vivid emotion in Row 4 of Figure \ref{fig:alfw_large}, and the mouths of our result in Row 3,5 have obviously better shape than 3DDFA \cite{dataset_aflw20003D_300WLP_zhu2016face}, RingNet \cite{unsuper_mul_sanyal2019learning_ring}, PRN \cite{super_fit_volu_exp_feng2018joint}, and Deng \etal \cite{unsuper_mul_ng2019accurate}. Besides, the face alignment results from Row 3 to Row 6 support that we achieve better face pose estimation, especially in large and extreme pose cases.


\noindent
\textbf{Texture, illumination shadings} We also visualize our result under geometry, texture, illumination shadings, and notice that our approach performs better than Tewari18 \etal \cite{unsuper_tewari2018self} and Tewari19 \etal \cite{unsuper_mul_tewari2019fml}, where the overlay result is very similar to the input image as Figure \ref{fig:exp_fml_5}. 
Further result and analysis about the result can be found in the \textit{suppl.} material.

MGCNet does not focus on the appearance of 3DMM as \cite{unsuper_tewari2018self,unsuper_mul_tewari2019fml,unsuper_tran2019towards,super_chen2019photo,super_zeng2019df2net,super_review3_galteri2019deep,super_fit_tran2018extreme,super_iter_sela2017unrestricted}, which is only constrained by render loss.
However, our multi-view geometry supervision can help render loss maximize the potential during training by accurate face alignment and depth value estimation.
This makes MGCNet able to handle 3DMM texture and illumination robustly.

\begin{table}
	\caption{(a)Performance comparison on AFLW2000-3D (68 landmarks). The normalized mean error (NME) for 2D landmarks with different yaw angles is reported. The first best result is highlighted in bold. (b) Average and standard deviation root mean squared error (RMSE) with \textit{mm} in three environments of MICC Florence.}
	\subtable[]{
		\centering
		\resizebox{0.445\textwidth}{!}{ 
			\begin{tabular}{|c|c|c|c|c|}
				\hlineB{3}
				\multicolumn{1}{|c||}{\textbf{Method}}
				& \multicolumn{1}{c|}{\textbf{0 to 30}}
				& \multicolumn{1}{c|}{\textbf{30 to 60}}
				& \multicolumn{1}{c|}{\textbf{60 to 90}}
				& \multicolumn{1}{c|}{\textbf{Mean}} \\
				\hlineB{3}
				3DDFA \cite{dataset_aflw20003D_300WLP_zhu2016face}      & 3.78     & 4.54   & 7.93       & 5.42
				\\
				3D-FAN \cite{dataset_bulat2017far_lsw3d}      & 3.61    & 4.34     & 6.87   & 4.94
				\\ 
				3DSTN \cite{super_fit_stn_bhagavatula2017faster}   &3.15     & 4.33     & 5.98   & 4.49 			
				\\ 
				CMD \cite{unsuper_zhou2019dense}   & -    & -     & -   & 3.98 
				\\ 
				PRN \cite{super_fit_volu_exp_feng2018joint}   &2.75     & 3.55     & 5.11   & 3.62 
				\\
				\hlineB{3}
				Ours+BL      & \textbf{2.75}     & \textbf{3.28}     & \textbf{4.31}      & \textbf{3.45}
				\\ 
				Ours+MGCNet      & \textbf{2.72}     & \textbf{3.12}     & \textbf{3.76}      & \textbf{3.20}
				\\ 
				\hlineB{3}
			\end{tabular}
		}
		\label{tab:aflw2D}
	} 
	\quad
	\subtable[]{
		\centering
		\resizebox{0.495\textwidth}{!}{ 
			\begin{tabular}{|c|c|c|c|}
				\hlineB{3}
				\multicolumn{1}{|c||}{\textbf{Method}}
				& \multicolumn{1}{c|}{\textbf{Cooperative}}
				& \multicolumn{1}{c|}{\textbf{Indoor}}
				& \multicolumn{1}{c|}{\textbf{Outdoor}} \\
				
				\hlineB{3}
				Zhu \etal \cite{dataset_aflw20003D_300WLP_zhu2016face}  & 2.69 $\pm$ 0.64     & 2.23 $\pm$ 0.49   & 2.22 $\pm$ 0.56
				\\
				Sanyal \etal \cite{unsuper_mul_sanyal2019learning_ring}  & 2.33 $\pm$ 0.43     & 2.19 $\pm$ 0.43   & 2.07 $\pm$ 0.45
				\\
				Feng \etal \cite{super_fit_volu_exp_feng2018joint}  & 2.30 $\pm$ 0.54     & 2.02 $\pm$ 0.50   & 2.10 $\pm$ 0.60
				\\ 
				Tran \etal \cite{super_fit_endtoend_iter_tuan2017regressing}  & 2.00 $\pm$ 0.55     & 2.05 $\pm$ 0.51   & 1.95 $\pm$ 0.51
				\\ 
				Genova \etal \cite{unsuper_genova2018unsupervised}            & 1.87 $\pm$ 0.61     & 1.86 $\pm$ 0.60   & 1.87 $\pm$ 0.57 
				\\ 
				Deng \etal \cite{unsuper_mul_ng2019accurate}                  & 1.83 $\pm$ 0.59     & 1.78 $\pm$ 0.53   & 1.78 $\pm$ 0.59 
				\\ 
				Ours                                                          & \textbf{1.73 $\pm$ 0.48}     & \textbf{1.78 $\pm$ 0.47}   & \textbf{1.75 $\pm$ 0.47 }
				\\ 
				\hlineB{3}
			\end{tabular}
		}
		\label{tab:micc_video}
		
	}
\end{table}

\subsection{2D Face Alignment}
The quantitative comparison of 6Dof pose is not conducted due to different camera intrinsic assumptions of different methods. Therefore, to validate that our MGCNet can mitigate the ambiguity of monocular face pose estimation, we evaluate our method on AFLW2000-3D, and compare our result with Zhu \etal \cite{dataset_aflw20003D_300WLP_zhu2016face} (3DDFA), Bulat and Tzimiropoulos  \cite{dataset_bulat2017far_lsw3d} (3D-FAN), Bhagavatula \etal \cite{super_fit_stn_bhagavatula2017faster} (3DSTN), Zhou \etal \cite{unsuper_zhou2019dense} (CMD), and Feng \etal \cite{super_fit_volu_exp_feng2018joint} (PRN). 
Normalized mean error (NME) is used as the evaluation metric, and the bounding box size of ground truth landmarks is deemed as the normalization factor. 
As shown in Table 1(a) Column 5, our result outperforms the best method with a large margin of $ \mathbf{12\%} $ improvement.
Qualitative results can be found in the \textit{suppl.} material.

Learning face pose from 2D features of monocular images leads to face pose ambiguity, the results of large and extreme face pose test samples suffer from this heavily. As the supervision of large and extreme face pose case is even less, which is not enough for training monocular face pose regressor. 
Our MGCNet provides further robust and dense supervision by multi-view geometry for face alignment in both frontal and profile face pose situations.
The comparison in Table 1(a) corroborates our point that the compared methods \cite{dataset_aflw20003D_300WLP_zhu2016face,dataset_bulat2017far_lsw3d,super_fit_stn_bhagavatula2017faster,unsuper_zhou2019dense,super_fit_volu_exp_feng2018joint} obviously degrades when the yaw angles increase from $ (30, 60) $ to $ (60, 90) $ in Column 4 of Table 1(a).
We also conduct an ablation study that our MGCNet outperforms the baseline, especially on large and extreme pose case.

\subsection{3D Face Reconstruction}
\textbf{MICC Florence with Video}
MICC Florence provides videos of each subject in cooperative, indoor and outdoor scenarios. For a fair comparison with Genova \etal \cite{unsuper_genova2018unsupervised}, Trans \etal \cite{super_fit_endtoend_iter_tuan2017regressing} and Deng \etal \cite{unsuper_mul_ng2019accurate}, we calculate error with the average shape for each video in different scenarios.
Following \cite{unsuper_genova2018unsupervised}, we crop the ground truth mesh to 95\textit{mm} around the nose tip and run iterative closest point (ICP) algorithm for rigid alignment.
The results of \cite{super_fit_endtoend_iter_tuan2017regressing} only contain part of the forehead region. For a fair comparison, we process the ground-truth meshes similarly. We use the point-to-plane root mean squared error(RMSE) as the evaluation metric.
We compare with the methods of Zhu \etal \cite{dataset_aflw20003D_300WLP_zhu2016face} (3DDFA), Sanyal \etal \cite{unsuper_mul_sanyal2019learning_ring} (RingNet), Feng \etal \cite{super_fit_volu_exp_feng2018joint} (PRN), Genova \etal \cite{unsuper_genova2018unsupervised}, Trans \etal \cite{super_fit_endtoend_iter_tuan2017regressing} and Deng \etal \cite{unsuper_mul_ng2019accurate}. 
Table 1(b) shows that our method outperforms state-of-the-art methods \cite{dataset_aflw20003D_300WLP_zhu2016face, unsuper_mul_sanyal2019learning_ring,super_fit_volu_exp_feng2018joint,unsuper_genova2018unsupervised,super_fit_endtoend_iter_tuan2017regressing,unsuper_mul_ng2019accurate} on all three scenarios.

\begin{figure}[htbp]
	\centering  
	\subfigure[]{
		\label{fig:exp_fml_5}
		\includegraphics[width=0.475\linewidth,scale=1.00]{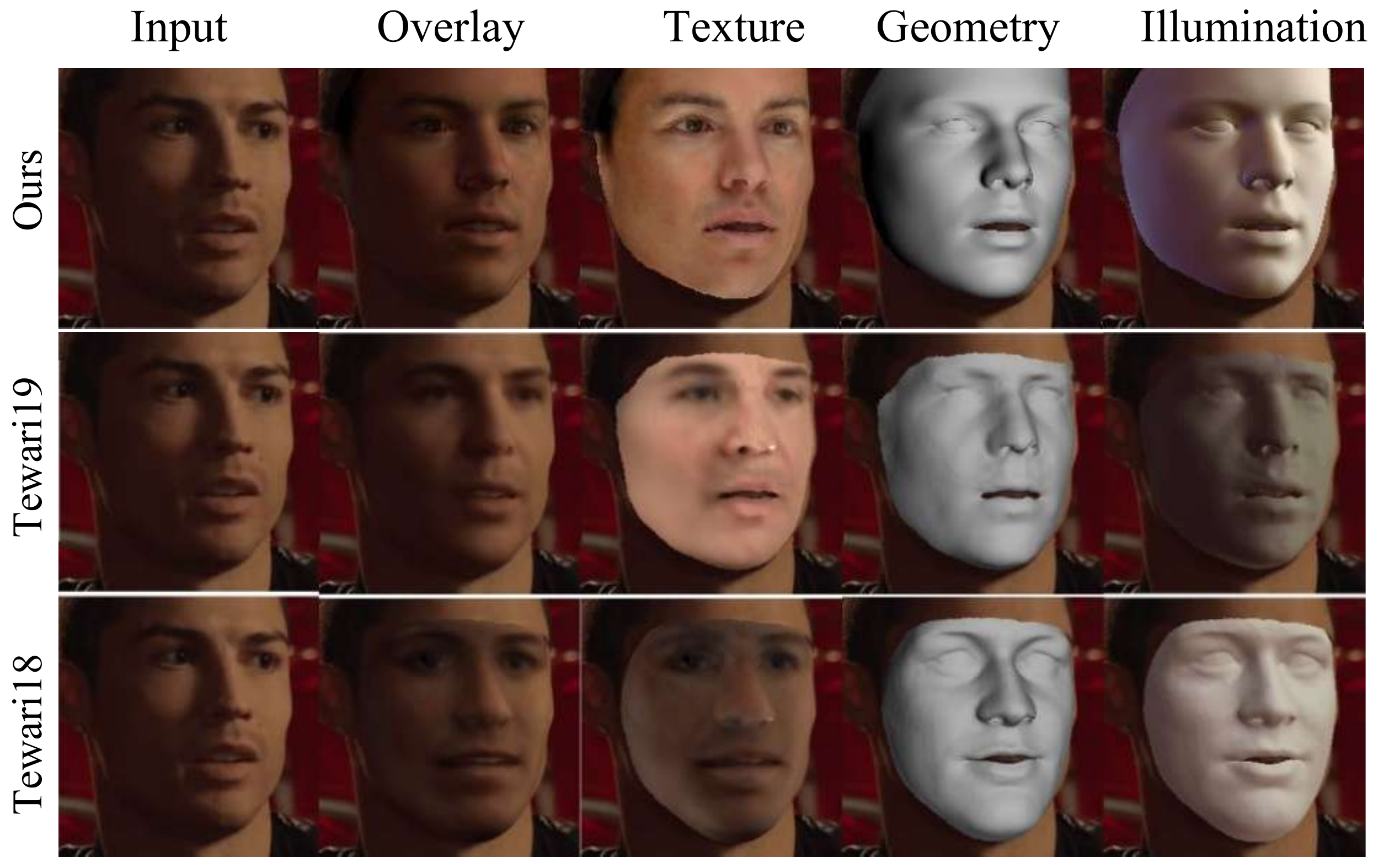}
	}
	\subfigure[]{
		\label{fig:eval_micc_render_error}
		\includegraphics[width=0.475\linewidth,scale=1.00]{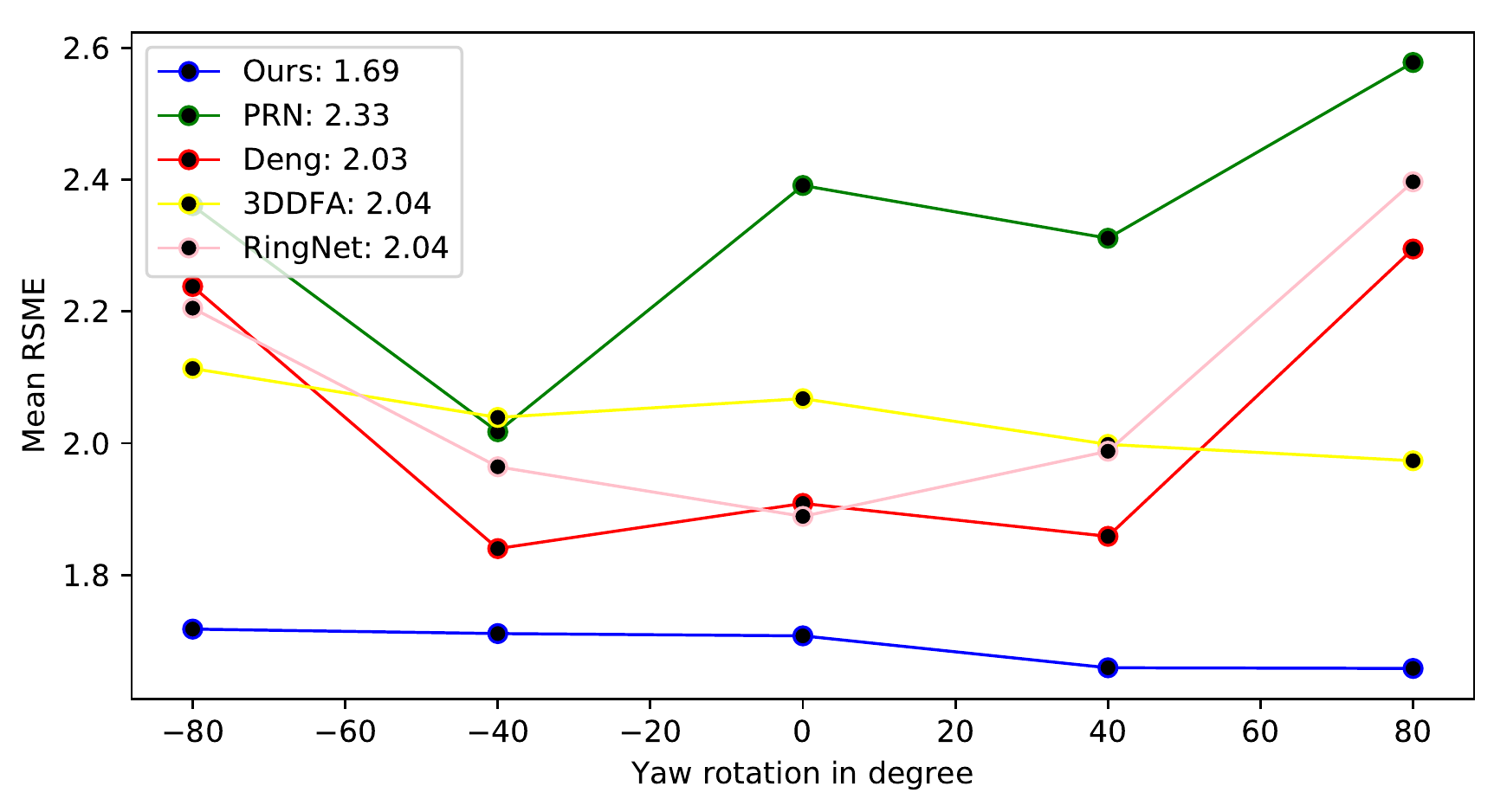}
	}
	\caption{(a)Comparison to Tewari18 \etal \cite{unsuper_tewari2018self} and Tewari19 \etal \cite{unsuper_mul_tewari2019fml}. Our MGCNet trained by multi-view consistency loss outperforms Tewari's results in face pose, illumination and geometry. Further result can be found in the \textit{suppl.} material. (b) Comparison with 3DDFA \cite{dataset_aflw20003D_300WLP_zhu2016face}, RingNet \cite{unsuper_mul_sanyal2019learning_ring}, PRN \cite{super_fit_volu_exp_feng2018joint}, and Deng \etal \cite{unsuper_mul_ng2019accurate} on MICC Florence rendered images. 
	}
\end{figure}

\noindent
\textbf{MICC Florence with Rendered Images}
Several current methods \cite{unsuper_mul_ng2019accurate,super_fit_yi2019mmface,super_fit_volu_exp_feng2018joint,super_fit_volu_jackson2017large} also generate rendered images as test input.
Following \cite{unsuper_mul_ng2019accurate,super_fit_yi2019mmface,super_fit_volu_exp_feng2018joint,super_fit_volu_jackson2017large}, we render face images of each subject with 20 poses: a pitch of -15, 20 and 25 degrees, yaw angles of -80, -60, 0, 60 and 80 degrees, and 5 random poses. We use the point-to-plane RMSE as the evaluation metric, and we process the ground truth mesh as above.
Figure \ref{fig:eval_micc_render_error} shows that our method achieves a significant improvement of $ \mathbf{17\%} $ higher than the state-of-the-art methods. 

The plot also shows that our MGCNet performs obvious improvement on the extreme pose setting $ x-axis [-80, 80] $ in Figure \ref{fig:eval_micc_render_error}. As we mitigate both pose and depth ambiguity by multi-view geometry consistency in the training process. Extreme pose sample benefits from this more significantly, since the extreme pose input images have even less 2D features. Profile face case contains more pronounced depth info (eg. bridge of the nose), where large error happen.

\noindent
\textbf{FRGC v2.0 Dataset}
FRGC v2.0 is a large-scale benchmark includes 4007 scans. We random pick 1335 scans as test samples, then we crop the ground truth mesh to 95\textit{mm} around the nose tip. We first use 3D landmark as correspondence to align the predict and ground truth result, then ICP algorithm is used as fine alignment. Finally, point-to-point mean average error (MAE) is used as the evaluation metric.
We compare with the methods of Galteri \etal \cite{super_review3_galteri2019deep} (D3R), 3DDFA \cite{dataset_aflw20003D_300WLP_zhu2016face}, RingNet \cite{unsuper_mul_sanyal2019learning_ring}, PRN \cite{super_fit_volu_exp_feng2018joint}, and Deng \etal \cite{unsuper_mul_ng2019accurate}. 
Table \ref{tab:frgc} shows that our method outperforms these state-of-the-art methods. Our MGCNet performs higher fidelity and accurate result on both frontal and profile face pose view.

\begin{table}
	\caption{(a) Comparison with D3R \cite{super_review3_galteri2019deep}, 3DDFA \cite{dataset_aflw20003D_300WLP_zhu2016face}, RingNet \cite{unsuper_mul_sanyal2019learning_ring}, PRN \cite{super_fit_volu_exp_feng2018joint}, and Deng \etal \cite{unsuper_mul_ng2019accurate} with MAE of \textit{mm} on FRGC v2.0 dataset. (b) Mean and standard deviation point-to-point RMSE with \textit{mm} on the BU-3DFE dataset \cite{dataset_bu3dfe_yin20063d,dataset_bu4dfe_yin20063d} compared with Tewari17 \etal \cite{unsuper_tewari2017mofa}, Tewari18 \etal \cite{unsuper_tewari2018self}, Tewari19 \etal \cite{unsuper_mul_tewari2019fml}, Deng \etal \cite{unsuper_mul_ng2019accurate}.
	}
	\centering
	\subtable[]{
		\centering
		\resizebox{0.415\textwidth}{!}{ 
			\begin{tabular}{c|c|c|c|c|c|c}
				\hlineB{3}
				\multicolumn{1}{c||}{\textbf{Method}}
				& \multicolumn{1}{c|}{\cite{super_review3_galteri2019deep}}
				& \multicolumn{1}{c|}{\cite{super_fit_volu_exp_feng2018joint}}
				& \multicolumn{1}{c|}{\cite{unsuper_mul_sanyal2019learning_ring}}
				& \multicolumn{1}{c|}{\cite{dataset_aflw20003D_300WLP_zhu2016face}}
				& \multicolumn{1}{c|}{\cite{unsuper_mul_ng2019accurate}}
				& \multicolumn{1}{c}{\textbf{Ours}} \\
				\hlineB{3}
				MAE  & 3.63  & 2.33  & 2.22  & 2.21       & 2.18 &  \textbf{1.93} \\
				Time & - & 9.8ms & 2.7ms & 75.7ms     & 20ms &  20ms \\
				\hlineB{3}
			\end{tabular}
		}
		\label{tab:frgc}
	} 
	\quad
	\subtable[]{
		\resizebox{0.45\textwidth}{!}{ 
			\begin{tabular}{c|c|c|c|c|c|c}
				\hlineB{3}
				\multicolumn{1}{c||}{\textbf{Method}}
				& \multicolumn{1}{c|}{\cite{unsuper_tewari2017mofa}}
				& \multicolumn{1}{c|}{\cite{unsuper_tewari2018self} Fine}
				& \multicolumn{1}{c|}{\cite{unsuper_tewari2018self} Coarse}
				& \multicolumn{1}{c|}{\cite{unsuper_mul_tewari2019fml}}
				& \multicolumn{1}{c|}{\cite{unsuper_mul_ng2019accurate}}
				& \multicolumn{1}{c}{\textbf{Ours}} \\
				\hlineB{3}
				Mean  & 3.22 & 1.83     & 1.81     & 1.79 & 1.63 & \textbf{1.55} \\
				Std   & 0.77 & 0.39     & 0.47     & 0.45 & 0.33 & \textbf{0.31} \\
				\hlineB{3}
			\end{tabular}
		}
		\label{tab:bu3dfe}
		
	}
\end{table}
\begin{table}[htbp]
	\caption{Evaluation of different training loss configurations.
		The ablation study performance of MGCNet is evaluated on MICC Florence 3D Face dataset \cite{dataset_florence} by RMSE. 
	}
	\centering
	\resizebox{0.85\textwidth}{!}{ 
		\begin{tabular}{c|c|c|c|c||c|c|c}
			\hlineB{3}
			\multicolumn{5}{c||}{\textbf{Loss Configuration}}  
			& \multicolumn{3}{c}{\textbf{MICC Florence video}}      \\                                                        
			
			\hline
			2D feature & Pixel Consistency & Depth Consistency & Epipolar & Covisible map
			
			& \multicolumn{1}{c|}{\cellcolor{red!20}Cooperative} 
			& \multicolumn{1}{c|}{\cellcolor{red!20}Indoor} 
			& \multicolumn{1}{c}{\cellcolor{red!20}Outdoor} 
			\\ 
			
			\hlineB{3}
			$\checkmark$  & - & - & - & - & 1.83   &  1.82  &  1.81    
			\\ 
			$\checkmark$  & $\checkmark$  & - & - & $\checkmark$ &  1.80 &  1.80 & 1.80                        
			\\ 
			$\checkmark$  & -& $\checkmark$  & - & $\checkmark$ &  1.77 &  1.79 & 1.80                         
			\\ 
			$\checkmark$  & - & -  & $\checkmark$ & - &  1.79 &  1.81  & 1.77  
			\\ 
			$\checkmark$ & $\checkmark$ & $\checkmark$ & - & $\checkmark$ &  1.76 &  1.81 & 1.81    
			\\ 
			$\checkmark$ & $\checkmark$ & $\checkmark$ & $\checkmark$ & - &  1.80 & 1.81 &   1.82   
			\\ 
			$\checkmark$  & $\checkmark$ & $\checkmark$  & $\checkmark$	& $\checkmark$ &  \textbf{1.73} & \textbf{1.78} & \textbf{1.75}               
			\\ %
			\hlineB{3}
		\end{tabular}
	}
	\label{tab:ablation}
\end{table}

\noindent
\textbf{BU-3DFE Dataset}
We evaluate our method on the BU-3DFE dataset following \cite{unsuper_mul_tewari2019fml}. Following \cite{unsuper_mul_tewari2019fml}, a pre-computed dense correspondence map is used to calculate a similarity transformation from predict mesh to the original ground-truth 3D mesh, and help to calculate the point-to-point RMSE.
From Table \ref{tab:bu3dfe}, the reconstruction error of our method is lower than the current state-of-art methods \cite{unsuper_tewari2017mofa,unsuper_mul_tewari2019fml,unsuper_tewari2018self,unsuper_mul_ng2019accurate}. 
Our MGCNet achieves better performance by using the multi-view geometry consistency loss functions in the training phase. Qualitative results can be found in the \textit{suppl.} material.

\subsection{Ablation study} \label{sec_abla}
To validate the efficiency of our multi-view geometry consistency loss functions.
We conduct ablation studies for each component on the MICC Florence dataset \cite{dataset_florence}, as shown in Table \ref{tab:ablation}.
The ablation study mainly focuses on the proposed multi-view geometry consistency loss functions. 
Firstly, we deem the baseline method as the model trained with only 2D feature losses, as in Row 1.
Secondly, we add our pixel consistency loss, dense depth consistency loss, and epipolar loss to the baseline in Row 2, Row 3 and Row 4.
It shows that these losses help produce lower reconstruction errors than the baseline, even when they are used separately.
Thirdly, comparing from Row 5 to Row 7, we combine multiple multi-view geometry loss functions and achieve state-of-the-art results, which demonstrates the effectiveness of the proposed self-supervised learning pipeline.
Finally, comparing from Row 6 to Row 7, we prove that our novel covisible map to solve self-occlusion in view synthesis algorithm can help training a more accurate model.
The qualitative ablation study is in the \textit{suppl.} material.

\section{Conclusion}
We have presented a self-supervised pipeline MGCNet for monocular 3D Face reconstruction and demonstrated the advantages of exploiting multi-view geometry consistency to provide more reliable constraint on face pose and depth estimation.
We emphasize on the occlusion-aware view synthesis and multi-view losses to make the result more robust and reliable. 
Our MGCNet profoundly reveals the capability of multi-view geometry consistency self-supervised learning in capturing both high-level cues and feature correspondences with geometry reasoning.
The results compared to other methods indicate that our MGCNet can achieve the outstanding result without costly labeled data.
Our further investigations will focus on multi-view or video-based 3D face reconstruction.
\section{Acknowledgements}
This work is supported by Hong Kong RGC GRF 16206819 \& 16203518 and T22-603/15N.
\bibliographystyle{splncs04}
\bibliography{dl_bib,tra_bib}
\let\cleardoublepage\clearpage

\pagestyle{headings}
\mainmatter
\def\ECCVSubNumber{2293}  

\title{Supplemental Material for \\ "Self-Supervised Monocular 3D Face Reconstruction by Occlusion-Aware Multi-view Geometry Consistency"} 

\titlerunning{MGCNet}
%
\author{Jiaxiang Shang\inst{1}\orcidID{0000-0001-7161-9765} \and Tianwei Shen\inst{1}\orcidID{0000-0002-3290-2258} \and
	Shiwei li\inst{1}\orcidID{0000-0003-0712-0059} \and Lei Zhou\inst{1}\orcidID{0000-0003-4988-5084} \\ \and 
	Mingmin Zhen\inst{1}\orcidID{0000-0002-8180-1023} \and Tian Fang\inst{2}\orcidID{0000-0002-5871-3455} \and \\ 
	Long Quan\inst{1}\orcidID{00000001-8148-1771}}
\authorrunning{J. Shang et al.}
%
\institute{Hong Kong University of Science and Technology \email{\{jshang,tshenaa,lzhouai,mzhen,quan\}@cse.ust.hk} \\
	\and Everest Innovation Technology \\ 
	\email{\{sli,fangtian\}@altizure.com}}
\maketitle

\section{Overview}
This supplementary document provides detailed evaluation results that are supplementary to the main paper.
We propose a self-supervised Multi-view Geometry Consistency based 3D Face Reconstruction framework (MGCNet), which helps mitigate the monocular face pose and depth ambiguity.
Firstly, we propose the detailed data pre-process pipeline in Section \ref{sec_data_pre}, then we introduce the quantitative evaluation datasets in Section \ref{sec_eval_dataset}.
Secondly, we introduce the morphable model and highlight that our MGCNet is a general framework in Section \ref{sec_3dmm}.
Thirdly, we evaluate the quantitative result by render error between the input image and rendered image in Section \ref{eval_celebA} and we show the qualitative ablation study in Section \ref{eval_abl}.
Furthermore, we show further comparison with Tewari19 \cite{unsuper_mul_tewari2019fml} under geometry, texture and lighting in Section \ref{compare_fml}, then we conduct further comparison with some methods on the in the wild images in Section \ref{compare_in_wild}. 
Finally, we demonstrate the qualitative comparisons against other methods on MICC Florence dataset \cite{dataset_florence} in Section \ref{compare_micc} and we also demonstrate some results from AFLW20003D \cite{dataset_aflw20003D_300WLP_zhu2016face} in Section \ref{compare_aflw}, which further certify our MGCNet performs accurate result on face alignment task. Then we show the qualitative result on BU-3DFE dataset \cite{dataset_bu3dfe_yin20063d,dataset_bu4dfe_yin20063d} in Section \ref{compare_bu3dfe}.
%
\section{Data Preprocess}  \label{sec_data_pre}
The images are automatically annotated by the 2D landmark detection method in \cite{dataset_bulat2017far_lsw3d} and the face detection method in \cite{zhang2017s3fd}.
We filter the face pose, face attribution, low-resolution images, and blurred images and obtain $\sim$390K face images from the above four datasets as our training set.
The images are scaled to a resolution of 224 $ \times $ 224. 

The multi-view images of the training dataset are captured with a consistent lighting condition across views. Theoretically, the photometric consistency will be violated if the lighting across views is dramatically different. Our MGCNet shares the same property with multi-view stereo methods that require overlap across views, this also the reason that we only use $ N=3 $ views in practice.

\section{Quantitative Evaluation Dataset}  \label{sec_eval_dataset}

\textbf{AFLW20003D} is constructed to evaluate face alignment on challenging in the wild images. This database contains the first 2000 images from AFLW \cite{dataset_alfw} with landmarks annotations. We use this database to evaluate the performance of our method on face alignment tasks \cite{dataset_aflw20003D_300WLP_zhu2016face}.

\noindent
\textbf{MICC Florence} is a 3D face dataset that contains 53 faces with their ground truth 
High-resolution 3D scans of human faces are acquired from a structured-light scanning system from each subject with several video sequences of varying resolution, conditions and zoom level \cite{dataset_florence}.

\noindent
\textbf{FRGC v2.0} includes 4007 scans of 466 individuals acquired with the frontal view from the shoulder level, with very tiny pose variations. 
About $ 60 \% $ of the faces have neutral expression, while the others show spontaneous expressions of disgust, happiness, sadness, and surprise  \cite{dataset_frgc}. 
Scans are given as matrices of 3D points of size $ [480, 640] $, with a binary mask indicating the valid points of the face (about 40 K on average). 

\noindent
\textbf{BU-3DFE} 
BU-3DFE database includes 100 subjects with 2500 facial expression models. 
The database presently contains 100 subjects (56$\%$ female, 44$\%$ male), ranging age from 18 years to 70 years old, with a variety of ethnic/racial ancestries, including White, Black, East-Asian, Middle-east Asian, Indian, and Hispanic Latino. Each subject performed seven expressions in front of the 3D face scanner. With the exception of the neutral expression, each of the six prototypic expressions (happiness, disgust, fear, angry, surprise and sadness) includes four levels of intensity \cite{dataset_bu3dfe_yin20063d,dataset_bu4dfe_yin20063d}. 

\section{3D Morphable Model}  \label{sec_3dmm}
Blanz and Vetter \cite{intro_3dmm_blanz1999morphable} introduce the 3D morphable model (3DMM). 
3DMM benefits the 3D face reconstruction by constraining the solution space, thereby simplifying the problem. 
In this paper, our goal is to estimate 3DMM parameters from a single photograph. 

We conduct our experiments with 3DMM model since it is still a general method that widely used by single-image based latest methods (as in works of \cite{unsuper_zhou2019dense,dataset_aflw20003D_300WLP_zhu2016face,unsuper_genova2018unsupervised,super_fit_endtoend_iter_tuan2017regressing,unsuper_mul_ng2019accurate,unsuper_tewari2017mofa}), and we have proved the superiority of our method over theirs under a fair comparison, as shown qualitatively in the main paper as well as our quantitative result.

As we have clarified in the main paper, our method is focused on improving single-view reconstruction quality via multi-view consistency, our proposed framework is general and is not limited to any specific face model. We believe other face models with better representation ability  \cite{unsuper_mul_sanyal2019learning_ring,unsuper_mul_tewari2019fml,unsuper_tewari2018self,unsuper_tran2018nonlinear,unsuper_tran2019towards} can easily plug into our proposed MGCNet.

\begin{figure*}[htb]
	\centering
	\includegraphics[width=0.875\linewidth,scale=1.00]{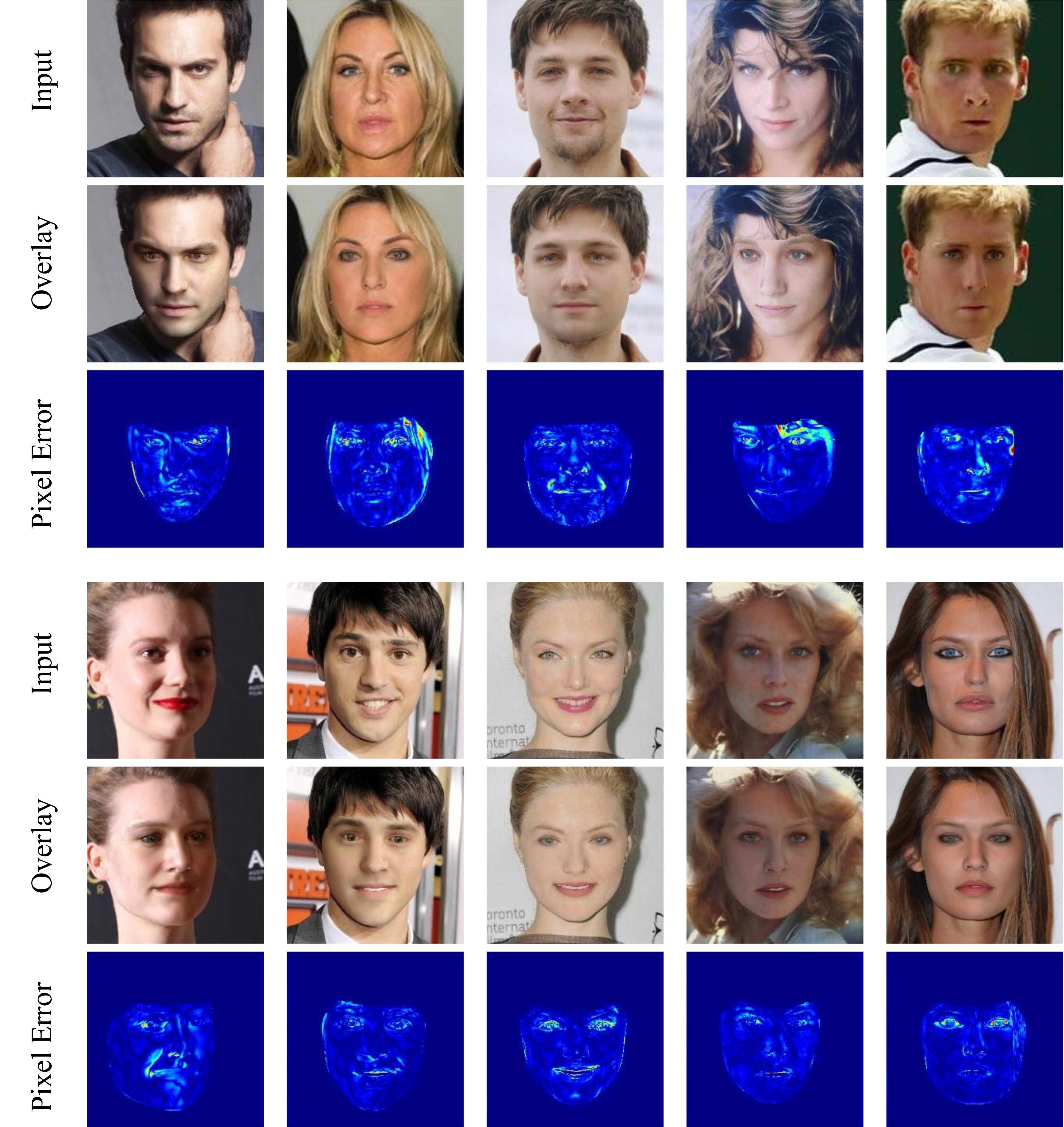}
	\caption{Quantitative evaluation of photometric error on the CelebA \cite{dataset_liu2015deep_celeba} dataset. The error map range is $ [0, 1] $
	}
	\label{fig:supp_celebA}
\end{figure*}

\section{Further Evaluation Result} 
\subsection{CelebA Dataset} \label{eval_celebA}
We evaluate the photometric error of our approach by heat maps on the CelebA dataset \cite{dataset_liu2015deep_celeba} as shown in Figure \ref{fig:supp_celebA} that these images are only used for testing and visualization. 
We achieve low pixel error, which benefits from using multi-view geometry consistency.
This also demonstrates better reconstruction capabilities of our MGCNet to in-the-wild images.

\subsection{Alation Study} \label{eval_abl}
\begin{figure*}[htb]
	\centering
	\includegraphics[width=0.875\linewidth,scale=1.00]{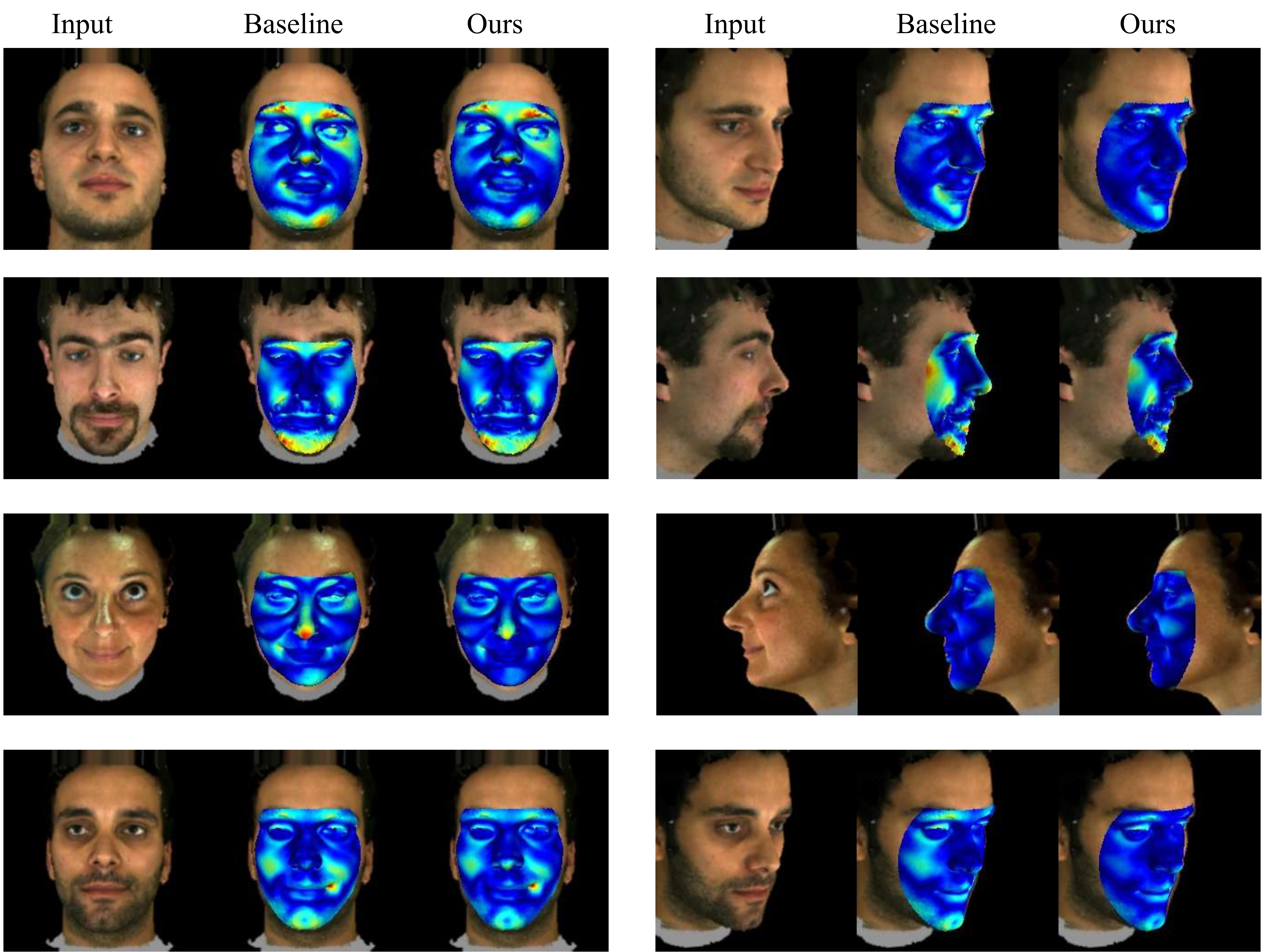}
	\caption{Quantitative evaluation of point-to-plane root mean squared error as the error map format on the  MICC Florence 3D Face dataset \cite{dataset_florence}. The error map range is $ [0, 8.29] $.
	}
	\label{fig:supp_abl_errMesh}
\end{figure*}

To evaluate the effects of multi-view geometry consistency losses on the quality of the reconstructed meshes.
We conduct ablation studies on the MICC Florence 3D Face dataset \cite{dataset_florence}, as shown in Figure \ref{fig:supp_abl_errMesh}. We calculate the point-to-plane root mean squared error, and normalize the error to a heatmap.
This heatmap indicate that the major improvements regions are jaw, nose and cheekbones region in frontal case, and eye contour, nose, cheekbones regions for the large-pose case.
Face geometry (especially in large pose cases), as well as better 3D pose estimation results, are the major improvements bring by our method, thanks to our multi-view geometry constraints that explicitly regularizes the geometry across different views.

\begin{figure*}[htb]
	\centering
	\includegraphics[width=0.875\linewidth,scale=1.00]{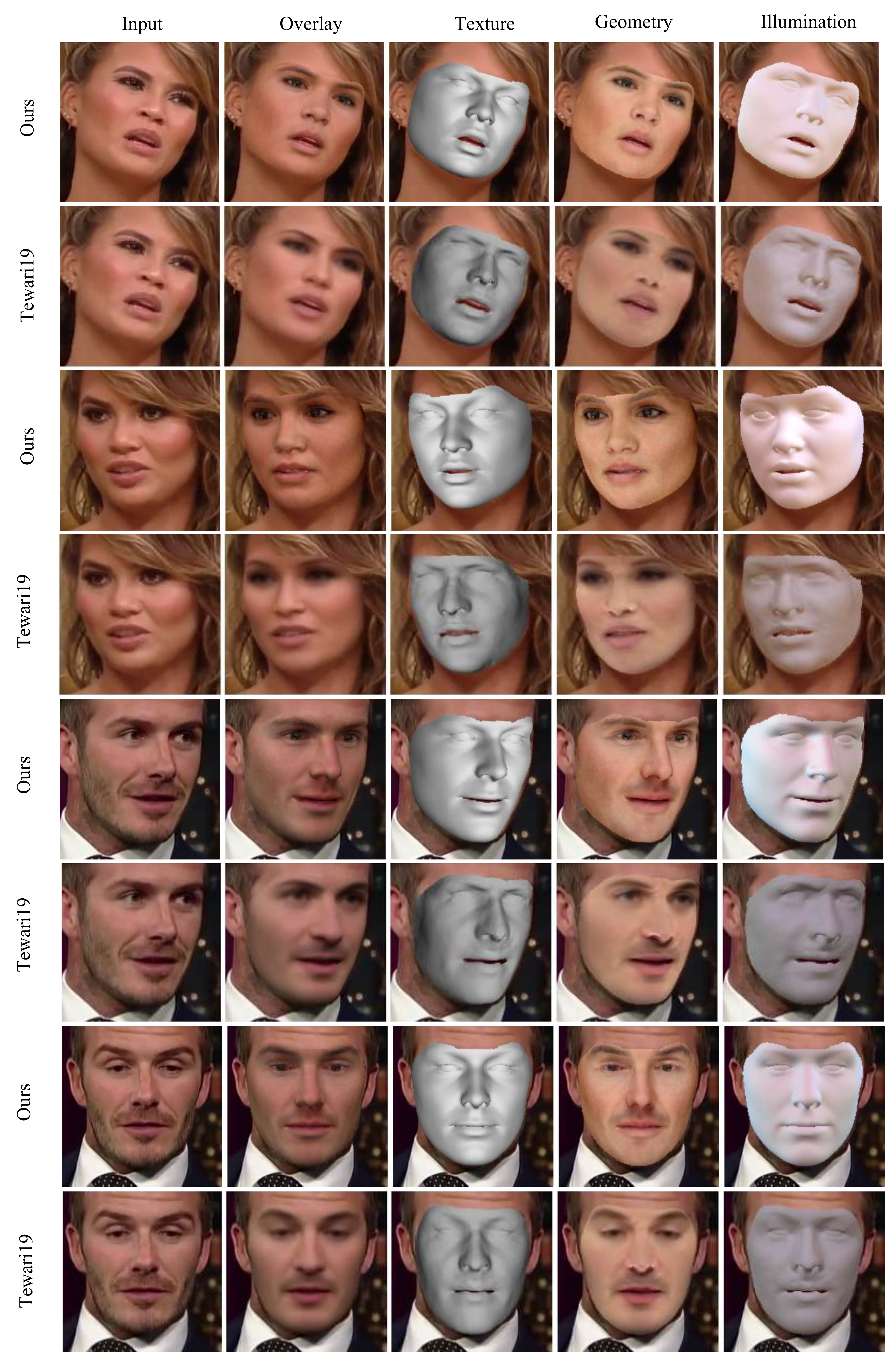}
	\caption{Quantitative evaluation compare with Tewari19 \cite{unsuper_mul_tewari2019fml}.
	}
	\label{fig:supp_fml_only19}
\end{figure*}

\subsection{Texture, illumination shadings} \label{compare_fml}
We compare our MGCNet with Tewari19 \cite{unsuper_mul_tewari2019fml} as Figure \ref{fig:supp_fml_only19}.
The result of Tewari19 \cite{unsuper_mul_tewari2019fml} results of geometry are “visually more detail” since they use a face representation more complicated than 3DMM used by our method. However, it is hard to say Tewari19 \cite{unsuper_mul_tewari2019fml} has better geometry results since our method does have a better quantitative result shown in the main paper.
The texture model used in Tewari19 \cite{unsuper_mul_tewari2019fml} is also different from 3DMM. Despite that, better geometry generated by our method leads to better texture via the render loss used, which can also support the validity of our MGCNet. As our method is focused on improving the reconstruction quality via multi-view consistency. Our MGCNet is a general system that is not limited to any specific face model.

\subsection{In the wild data} \label{compare_in_wild}
Secondly, we visualize our result under geometry overlay situation compared with Richardson \etal \cite{unsuper_richardson2017learning}, Sela \etal \cite{super_iter_sela2017unrestricted}, Tewari17 \etal \cite{unsuper_tewari2017mofa}, Tewari19 \etal \cite{unsuper_mul_tewari2019fml} and RingNet \etal \cite{unsuper_mul_sanyal2019learning_ring}, and we notice that our approach performs better than methods \cite{unsuper_richardson2017learning,super_iter_sela2017unrestricted,unsuper_tewari2017mofa,unsuper_mul_tewari2019fml,unsuper_mul_sanyal2019learning_ring} as shown in Figure \ref{fig:comp_inthewild}.

We also show some detail intermediate result about \textit{Figure 5 in the main paper} in Figure \ref{fig:aflwcelebA_fml}.

\subsection{MICC Florence Dataset} \label{compare_micc}
\begin{figure*}[htb]
	\centering
	\includegraphics[width=0.95\linewidth,scale=1.00]{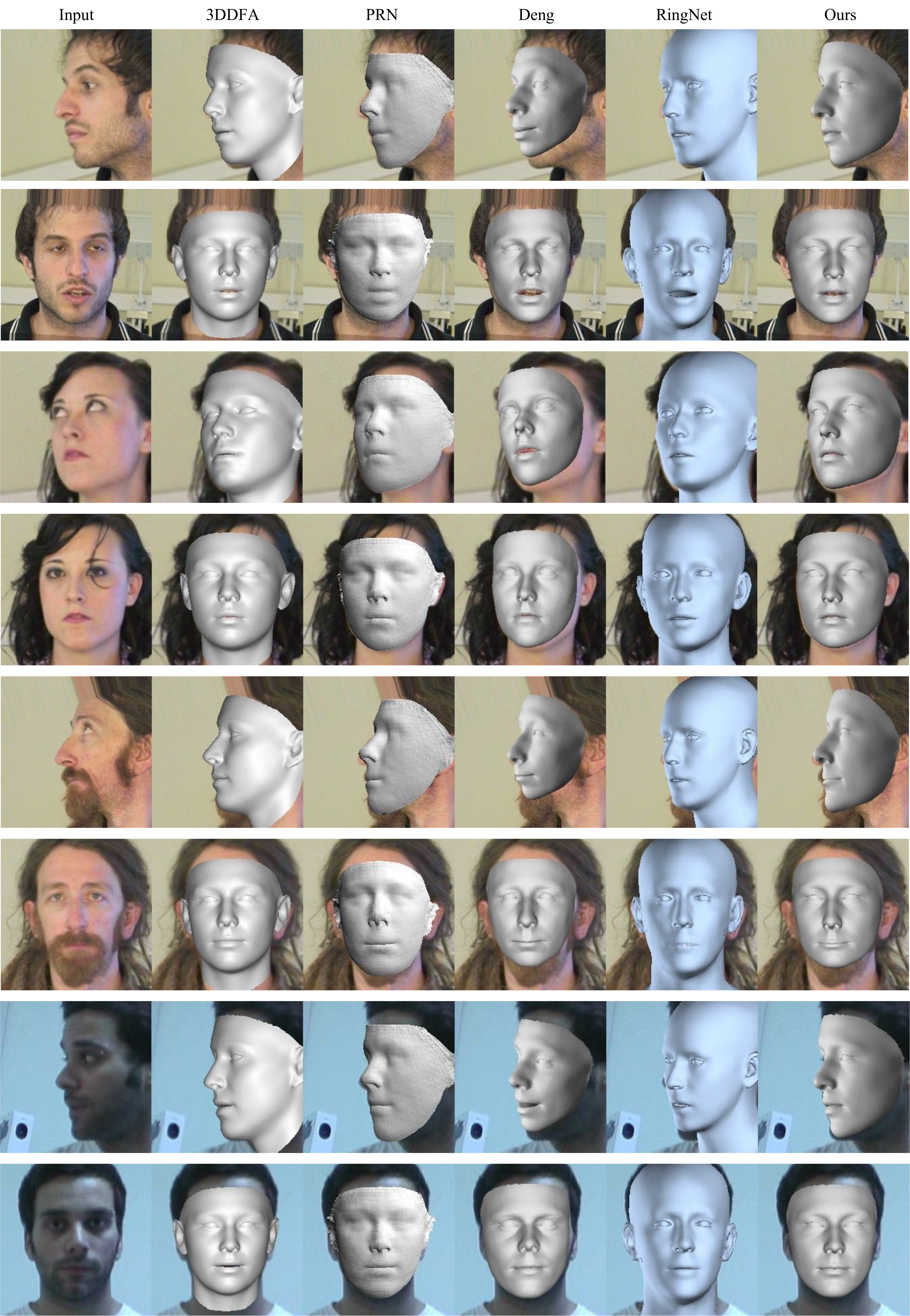}
	\caption{Comparison with Zhu \etal \cite{dataset_aflw20003D_300WLP_zhu2016face} (3DDFA), Sanyal \etal \cite{unsuper_mul_sanyal2019learning_ring} (RingNet), Feng \etal \cite{super_fit_volu_exp_feng2018joint} (PRN), and Deng \etal \cite{unsuper_mul_ng2019accurate}.}
	\label{fig:comp_micc}
\end{figure*}

\begin{figure*}[htbp]
	\centering
	\includegraphics[width=0.95\linewidth,scale=1.00]{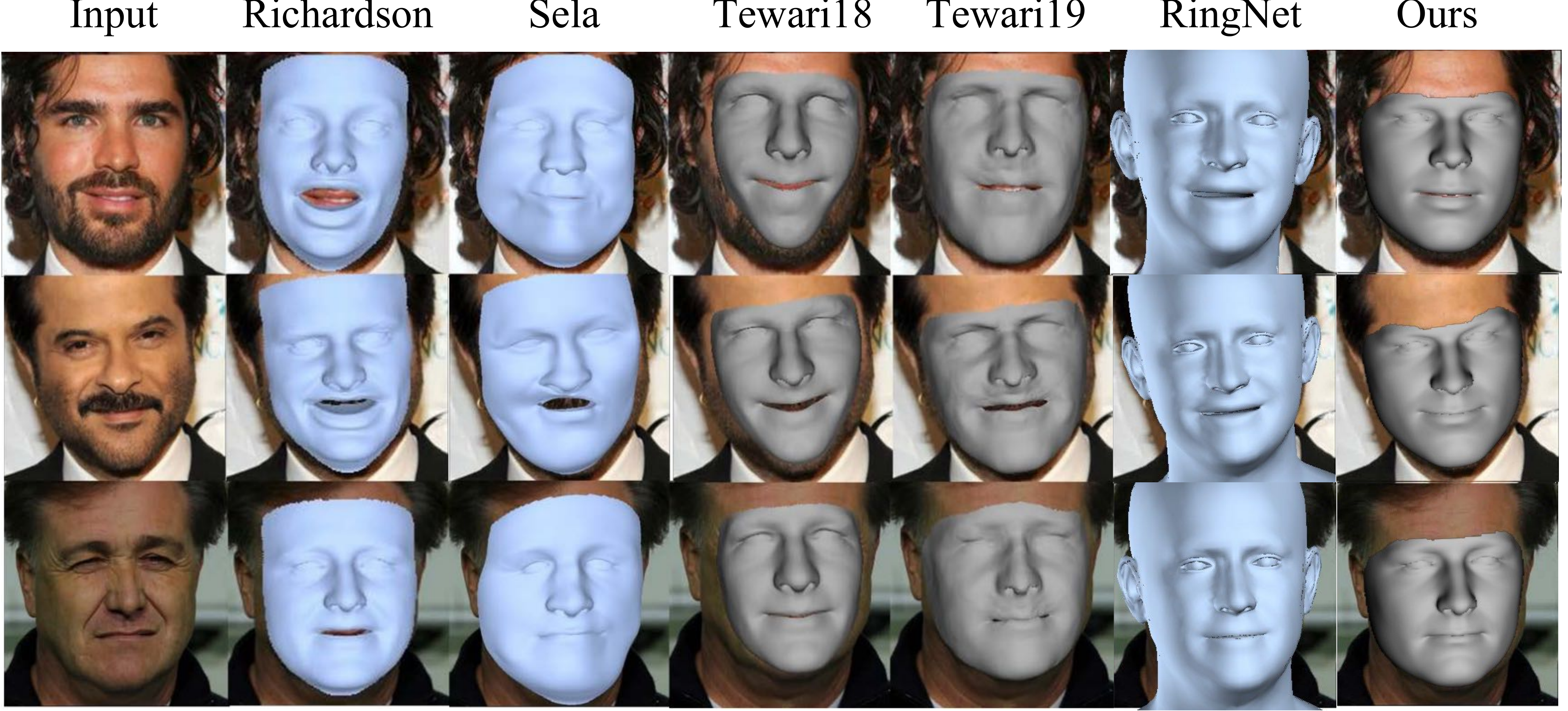}
	\caption{
		Comparison with Richardson \etal \cite{unsuper_richardson2017learning}, Sela \etal \cite{super_iter_sela2017unrestricted}, Tewari17 \etal \cite{unsuper_tewari2017mofa}, Tewari19 \etal \cite{unsuper_mul_tewari2019fml} and RingNet \etal \cite{unsuper_mul_sanyal2019learning_ring}.		
		Our MGCNet trained by multi-view consistency loss outperforms these state-of-the-art methods in face reconstruction geometry}
	\label{fig:comp_inthewild}
\end{figure*}

Firstly, we compare our MGCNet with Zhu \etal \cite{dataset_aflw20003D_300WLP_zhu2016face} (3DDFA), Sanyal \etal \cite{unsuper_mul_sanyal2019learning_ring} (RingNet), Feng \etal \cite{super_fit_volu_exp_feng2018joint} (PRN), and Deng \etal \cite{unsuper_mul_ng2019accurate}. For each sample in MICC Florence, we pick both front face images and large face pose images as test data. We show the geometry overlay of the reconstruction result, which we achieve more accurate results than the most methods, and we get better results than Deng \etal \cite{unsuper_mul_ng2019accurate} in the large pose case as Figure \ref{fig:comp_micc}.

\subsection{AFLW20003D Dataset} \label{compare_aflw}
AFLW20003D is constructed by \cite{dataset_aflw20003D_300WLP_zhu2016face} to evaluate face alignment. 
Since the images are captured in the wild and show large variations in pose and appearance, which is a challenging 3D face alignment dataset. We use this database to evaluate the performance of our method on face alignment tasks.

\begin{figure}[htbp]
	\centering
	\includegraphics[width=0.95\linewidth,scale=1.00]{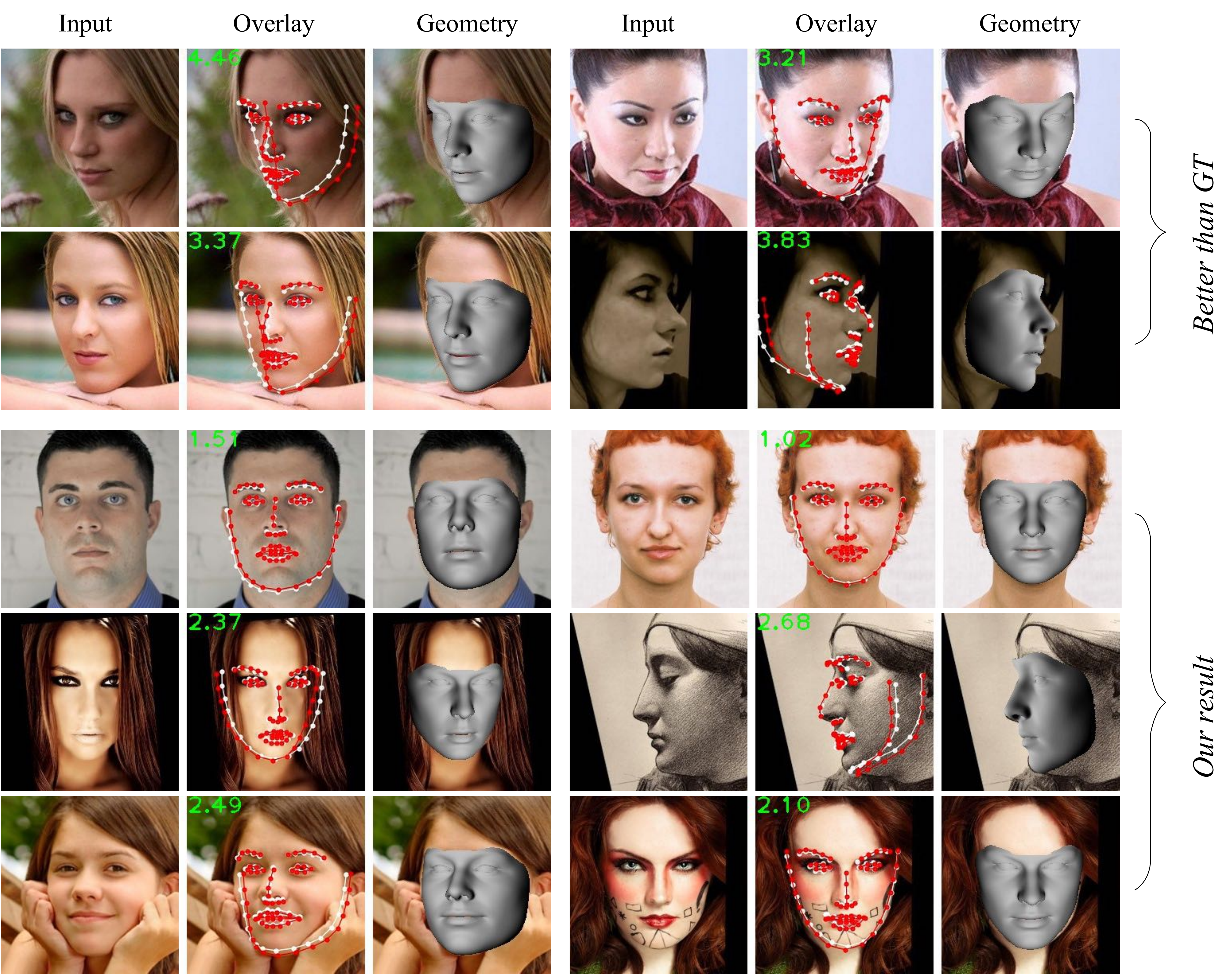}
	\caption{Examples from AFLW20003D dataset \cite{dataset_aflw20003D_300WLP_zhu2016face} show that our predictions are more accurate than ground truth in some cases. }
	\label{fig:aflw_better}
\end{figure}

\begin{figure}[htbp]
	\centering
	\includegraphics[width=0.95\linewidth,scale=1.00]{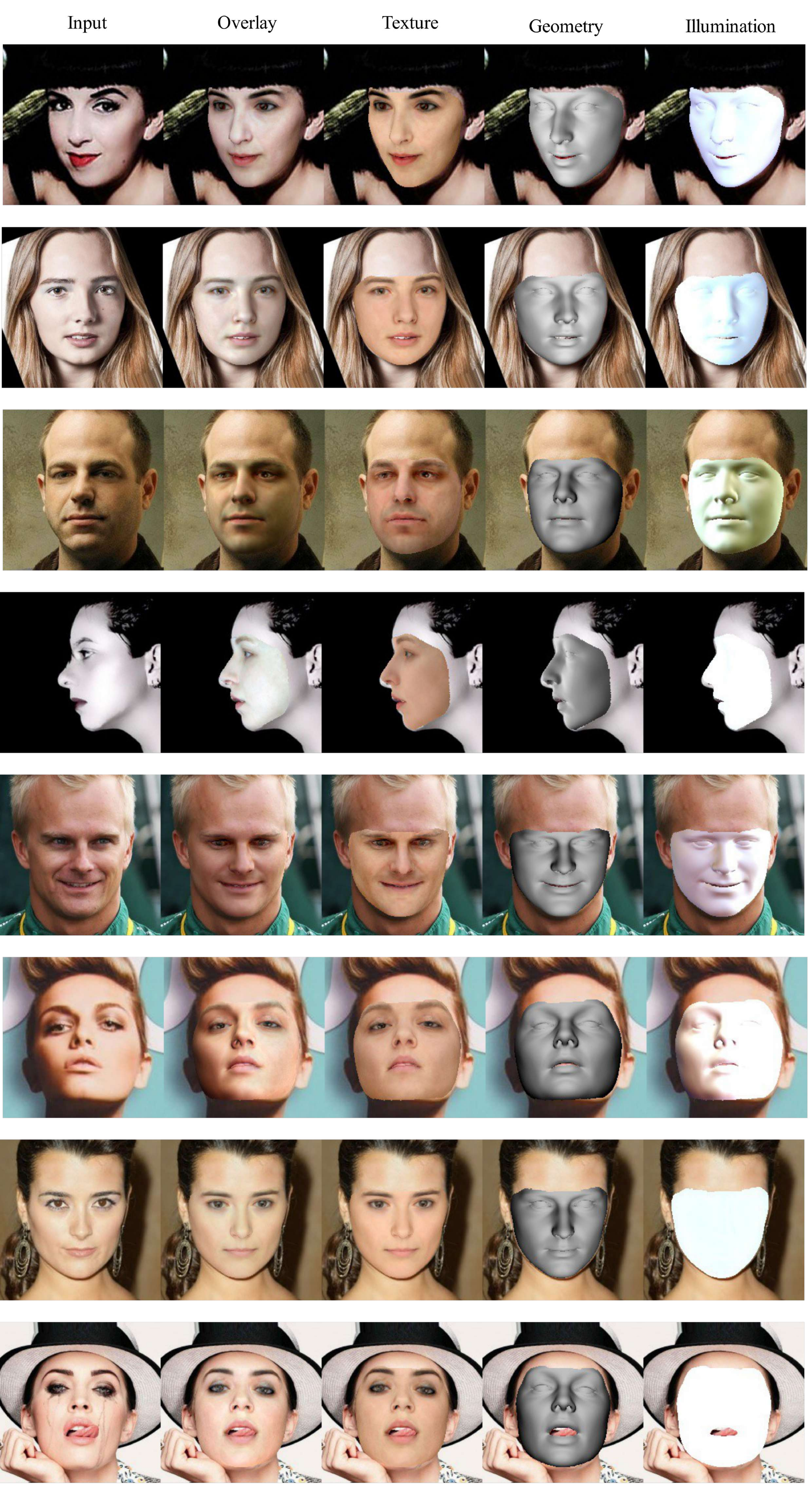}
	\caption{Face reconstruction results under texture, geometry and illumination of our method on AFLW20003D \cite{dataset_aflw20003D_300WLP_zhu2016face} and CelebA \cite{dataset_liu2015deep_celeba}. }
	\label{fig:aflwcelebA_fml}
\end{figure}

As a supplementary to the quantitative evaluation in the main paper, we first demonstrate some results even better than the ground truth from AFLW20003D \cite{dataset_aflw20003D_300WLP_zhu2016face} in Figure \ref{fig:aflw_better}. Besides, we also show our result that performs accurate face alignment results, where red lines are predicted landmarks by our method, white lines are ground truth from \cite{dataset_aflw20003D_300WLP_zhu2016face}. 

\begin{figure}[htbp]
	\centering
	\includegraphics[width=0.95\linewidth,scale=1.00]{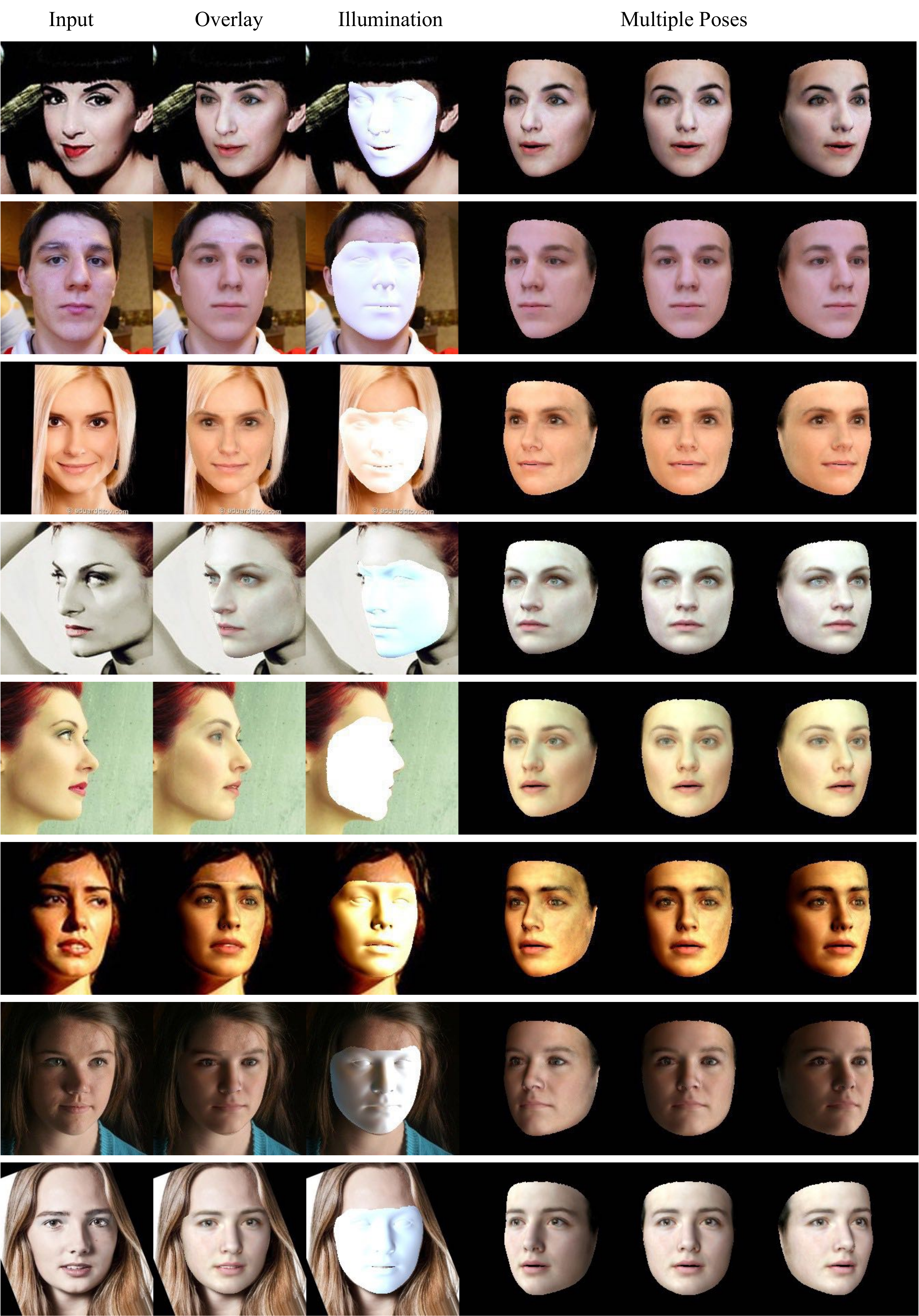}
	\caption{Face reconstruction results of our method on AFLW20003D \cite{dataset_aflw20003D_300WLP_zhu2016face} }
	\label{fig:aflw}
\end{figure}

Furthermore, we visual our MGCNet result from multiple viewpoints in Figure \ref{fig:aflw} on AFLW20003D \cite{dataset_aflw20003D_300WLP_zhu2016face}, which shows that we get vivid reconstruction results.

\subsection{BU-3DFE Dataset} \label{compare_bu3dfe}
\begin{figure*}[htbp]
	\centering
	\includegraphics[width=0.9\linewidth,scale=1.00]{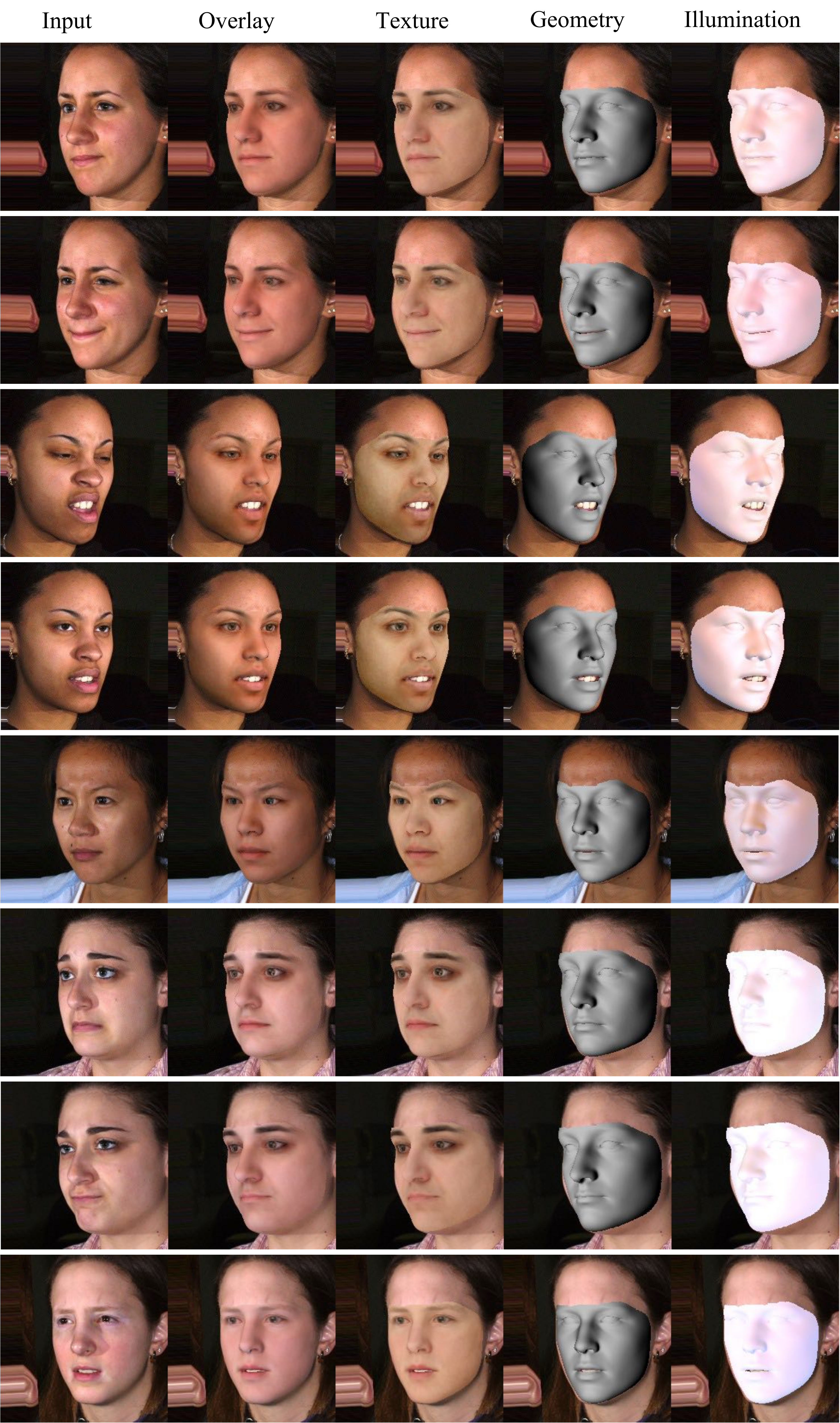}
	\caption{Our accurate result in face pose, texture, geometry and illumination on BU-3DFE dataset \cite{dataset_bu3dfe_yin20063d,dataset_bu4dfe_yin20063d}}
	\label{fig:bu3dfe}
\end{figure*}

We present more qualitative reconstruction results of our MGCNet on BU-3DFE dataset \cite{dataset_bu3dfe_yin20063d,dataset_bu4dfe_yin20063d}.
In Figure \ref{fig:bu3dfe}, we show six samples with various expressions, the reconstructed 3D face showed with face pose, texture, geometry and illumination.
This quantitative evaluation of our geometry reconstruction on the BU-3DFE dataset  \cite{dataset_bu3dfe_yin20063d,dataset_bu4dfe_yin20063d} shows that our MGCNet can even handle different expression situation. 
For the prediction of albedo and lighting, while the texture quality would be benefited from better geometry implicitly via the render loss, the ambiguity in illumination and face albedo is an intrinsic issue due to the problem’s nature, and our SH lighting is RGB-channel, limit the SH lighting to one channel will help.

\end{document}